\documentclass[a4paper,fleqn]{cas-sc}

\usepackage[numbers]{natbib}
\usepackage{amsmath,amssymb}
\usepackage{tabularx}
\usepackage{booktabs}
\usepackage{graphicx}
\graphicspath{{sections/}{./}}
\usepackage{xurl}
\usepackage{array}
\usepackage{capt-of}
\usepackage{float}
\usepackage{placeins}
\def\tsc#1{\csdef{#1}{\textsc{\lowercase{#1}}\xspace}}
\tsc{WGM}
\tsc{QE}
\tsc{EP}
\tsc{PMS}
\tsc{BEC}
\tsc{DE}

\begin{document}

\let\WriteBookmarks\relax

%
\setlength{\textfloatsep}{10pt plus 2pt minus 2pt}
\setlength{\floatsep}{8pt plus 2pt minus 2pt}
\setlength{\intextsep}{8pt plus 2pt minus 2pt}
\setcounter{topnumber}{2}
\setcounter{bottomnumber}{1}
\setcounter{totalnumber}{3}
\renewcommand{\topfraction}{0.95}
\renewcommand{\bottomfraction}{0.85}
\renewcommand{\textfraction}{0.05}
\renewcommand{\floatpagefraction}{0.85}


\title[mode=title]{CG-HAF: An Interpretable Global--Local Lesion-Burden Fusion Framework for Ordinal Acne Severity Grading in Agentic Skincare Support}

\shortauthors{M. M. Shahriar et al.}

\shorttitle{CG-HAF for Ordinal Acne Severity Grading}






\author[1]{Muhammad Muhtasim Shahriar}
[orcid=0009-0008-4954-3811]
\ead{shahriarmuhtasim@gmail.com}

\author[2]{Md. Naimur Asif Borno}
[orcid=0009-0006-1416-1771]
\ead{naimurborno@gmail.com}

\author[3]{Saad Aloteibi}
\ead{SaadAloteibi@ksu.edu.sa}

\author[4,5,6]{Mohammad Ali Moni}
[orcid=0000-0003-0756-1006]
\ead{m.moni@uq.edu.au}


\affiliation[1]{
    organization={Department of Computer Science, International Islamic University Chittagong (IIUC)},
    city={Chittagong},
    country={Bangladesh}
}

\affiliation[2]{
    organization={Department of Mechatronics, Rajshahi University of Engineering and Technology (RUET)},
    city={Rajshahi},
    country={Bangladesh}
}

\affiliation[3]{
    organization={Department of Computer Science and Engineering, College of Applied Studies, King Saud University},
    city={Riyadh},
    postcode={11437},
    country={Saudi Arabia}
}

\affiliation[4]{
    organization={AI \& Digital Health Technology, Rural Health Research Institute, Charles Sturt University},
    city={Orange},
    state={NSW},
    postcode={2800},
    country={Australia}
}

\affiliation[5]{
    organization={AI \& Digital Health Technology, Artificial Intelligence and Cyber Futures Institute, Charles Sturt University},
    city={Bathurst},
    state={NSW},
    postcode={2795},
    country={Australia}
}

\affiliation[6]{
    organization={School of Health and Rehabilitation Sciences, The University of Queensland},
    city={St Lucia, Brisbane},
    state={QLD},
    postcode={4072},
    country={Australia}
}


\begin{abstract}
Ordinal acne severity grading requires distinguishing visually similar neighboring grades while jointly weighing holistic facial appearance and localized lesion burden -- evidence that most existing approaches collapse into a single opaque representation. We introduce CG-HAF, a global--local fusion framework that instead keeps this evidence explicit: averaged holistic severity probabilities from independently trained classifiers are combined with structured lesion-burden descriptors from an object detector (lesion count, detection confidence, lesion area) into a compact representation, from which a lightweight, interpretable classifier produces the final grade. On a widely used benchmark, this fusion yields a clear, statistically supported improvement over global-evidence-only baselines, with the largest gains on the most severe cases. Testing on an independent dataset with a different grading standard shows that strong within-dataset performance does not transfer automatically, and a follow-up diagnostic attributes much of this gap to mismatched grading criteria rather than detection failure alone. These findings support interpretable global--local fusion as an effective strategy for ordinal acne grading while highlighting criterion alignment as key to cross-dataset portability, with a further illustration of how the resulting severity signal can support transparent, non-diagnostic decision-making in skincare applications.
\end{abstract}


\maketitle


\section{Introduction}
\label{sec:introduction}

Acne severity grading is a structured visual assessment task in which overall facial presentation and lesion burden jointly determine the assigned severity level. ACNE04 formalized this setting by pairing facial images with lesion-count information across four severity grades and showed that neighboring grades can share similar appearances \cite{ref01}. A recent systematic review characterizes conventional acne assessment as relying mainly on global severity grading and lesion counting, two sources of evidence that are complementary but individually subjective or labor-intensive \cite{ref02}. Because the target labels Mild, Moderate, Severe, and Very Severe form an ordered progression, the magnitude and direction of grading errors matter, and label-distribution formulations have been used to model ambiguity near grade boundaries rather than treating each image as an isolated class \cite{ref01}, \cite{ref03}. Evaluation should therefore capture ordinal agreement and class-specific behavior alongside accuracy.

Global and local evidence sources have complementary strengths. Global image-level models capture holistic cues such as erythema, texture, and overall lesion distribution, and global--local architectures such as KIEGLFN show that local skin information can complement whole-face representations \cite{ref04}. Diagnostic Evidence Distillation likewise exploits fine-grained lesion types and counts as training-time guidance, although its final student model still infers severity from the image alone \cite{ref05}. Detector- and count-centered pipelines instead keep local evidence explicit: AcneDet detects lesion types and passes their counts to a downstream grading model \cite{ref06}, and interpretable detection models can localize lesions and derive severity directly from detections \cite{ref07}. Local analysis remains difficult because lesions can be small, densely distributed, and sensitive to illumination and scale \cite{ref08}, and datasets are often strongly imbalanced at higher grades, obscuring minority Severe and Very Severe cases in aggregate metrics \cite{ref02}, \cite{ref06}. This complementarity motivates a fusion strategy that keeps holistic image evidence and explicit lesion burden simultaneously observable at the final decision stage.

Existing approaches, however, stop short of this goal. Label-distribution learning and severity-aware label smoothing address ordinal ambiguity \cite{ref01}, \cite{ref03}; ensembles such as AcneGrader improve robustness at the cost of maintaining multiple base models \cite{ref09}; global--local and prior-knowledge-guided methods combine whole-face and lesion-related cues \cite{ref04}, \cite{ref10}; and evidence-distillation and lesion-centric systems transfer or estimate severity from detected lesions \cite{ref05}, \cite{ref11}, \cite{ref06}, \cite{ref12}. Broader evidence also indicates that subtle localized cues and ordinal structure require dedicated evaluation \cite{ref13}, that grading protocols vary substantially across studies \cite{ref14}, and that the acne-AI literature is dominated by retrospective internal evaluation with limited external validation \cite{ref02}. None of these approaches establishes the incremental value of keeping image-level severity probabilities and detector-derived lesion-burden measurements jointly observable in a compact, interpretable final representation, nor does it examine this fusion through controlled feature ablation whose value is demonstrated beyond a single endpoint score.

To address this gap, we propose CG-HAF, a structured global--local evidence-fusion framework for ordinal acne severity grading. CG-HAF couples a ConvNeXtV2-Large global image branch with a YOLOv12s local lesion-evidence branch, aggregates severity probabilities across three independently trained global branches to reduce dependence on a single training run, and fuses them with count-derived grade indicators, normalized lesion counts, detection-confidence statistics, and lesion-area evidence into an interpretable 12-dimensional representation. A validation-fitted, balanced logistic-regression meta-classifier produces the final grade from this representation, and the evaluation separates fusion-feature ablation from training-component ablation while reporting ordinal agreement and high-severity recall alongside accuracy.

Acne grading criteria are not interchangeable, which raises a further, independent challenge for external transport: ACNE04 follows a Hayashi-aligned, count-oriented severity formulation, whereas PLSBRACNE01 follows a Pillsbury-based criterion in which lesion type and lesion burden jointly determine severity \cite{ref05}. We therefore separate two questions: whether the ACNE04-trained CG-HAF system transfers without adaptation to an independent PLSBRACNE01 cohort, evaluated through a label-blind, frozen cross-dataset and cross-criterion test; and whether target-criterion severity information remains recoverable from the same expert lesion evidence under leakage-controlled criterion adaptation, examined separately through an annotation-assisted nested cross-validation diagnostic.

The contributions of this work are as follows:
\begin{itemize}
\item A structured global--local evidence-fusion framework, CG-HAF, that combines holistic image-level severity probabilities with explicit detector-derived lesion evidence in a compact, interpretable 12-dimensional representation, with severity probabilities aggregated across three independently trained global branches to reduce dependence on any single stochastic training run.
\item A controlled evaluation protocol that isolates the effect of fusion-feature design from that of training components, reporting accuracy, class balance, ordinal agreement (QWK), grading error (MAE), calibration, and class-specific recall, together with a safety-constrained model-selection rule that rejects candidates showing reduced Severe or Very Severe sensitivity even when aggregate metrics improve.
\item A non-diagnostic skincare orchestration example illustrating how the structured CG-HAF outputs can support safety-grounded downstream decision-making.
\item An independent, label-blind, frozen cross-dataset and cross-criterion evaluation on PLSBRACNE01, together with a leakage-controlled, annotation-assisted diagnostic that separates the limitations of a fixed count-to-grade mapping from the added value of criterion-aware lesion-type evidence.
\end{itemize}
\FloatBarrier
\section{Related Work}
\label{sec:related-work}

\subsection{Ordinal Grading and Local Lesion Evidence}
\label{sec:ordinal-acne}

Acne severity prediction is an ordered multiclass problem in which the meaning of an error depends on its distance across grades. Systems that do not optimize an explicit ordinal objective often represent this structure indirectly through lesion burden. Zhang and Ma coupled a classification module that predicts severity and lesion count with a localization module, thereby exposing several severity-relevant outputs within a single ensemble rather than a single categorical label \cite{ref15}. A YOLOv11 study derived four severity categories from detected lesion counts using a Hayashi-style rule and showed that missed detections propagated directly into severity misclassification \cite{ref16}. ClearFace similarly localizes and classifies lesions before computing an Investigator's Global Assessment score from clinically weighted counts \cite{ref17}, while detector studies such as Deep Acne identify lesion categories and regions that can support, but do not independently encode, ordered whole-image severity \cite{ref18}.

Local lesion analysis has progressed from one-stage detectors that emphasize small-object sensitivity to multi-scale and efficiency-oriented designs. An enhanced YOLOv7 modified the backbone, attention, and anchors for dense facial acne on ACNE04 \cite{ref19}, and a controlled YOLOv8 study showed that optimizer choice alone affects precision, recall, and detection of subtle papules \cite{ref20}. Skin appearance also affects local detection: a YOLOv11 study reported persistent errors for low-contrast, underrepresented lesions across skin tones \cite{ref21}, and a parameter-reduced YOLOv11n generalized poorly to an external dataset despite competitive internal cross-validation \cite{ref22}. Earlier mobile-oriented YOLO research demonstrated real-time lesion-type recognition from DermNet-derived images without addressing ordered whole-face severity \cite{ref23}. Across these studies, localization, type, count, confidence, and spatial extent provide rich local evidence, but high detection accuracy alone does not establish reliable ordinal grading; local evidence must therefore be interpreted alongside global facial context.

\subsection{Global--Local and Evidence-Fusion Approaches}
\label{sec:global-local-fusion}

Hybrid grading methods differ mainly in \emph{where} global and local information is fused and whether local evidence remains observable in the final grade. Outside acne, Yoon et al. segment facial morphological features at high resolution and relate measured feature areas to severity, preserving a direct link between localized morphology and a downstream score \cite{ref24}. CNN--Transformer coupling in broader dermatology fuses convolutional locality with long-range context for lesion segmentation, but this fusion occurs in latent feature space rather than at an interpretable decision layer \cite{ref25}, and multimodal-dermatology reviews describe the resulting trade-off between combining complementary evidence and added data/validation burden \cite{ref26}. Latent fusion can capture complex interactions but is difficult to audit; structured evidence fusion instead supplies the final predictor with measurements that have identifiable semantic roles.

CG-HAF adopts the latter strategy. Its global stream contributes four image-level severity probabilities, aggregated across three independently trained runs. In contrast, its detector stream contributes a count-derived severity indicator, continuous lesion count, detection-confidence statistics, and lesion-area burden. A lightweight, balanced logistic regression meta-classifier combines these signals into a compact 12-dimensional detector-to-grading representation, so that local evidence explicitly enters the final decision rather than serving only as a training cue or a hidden embedding. This formulation also enables representation-level evaluation: feature groups can be introduced progressively to test whether count-derived, continuous-burden, confidence, and area evidence add information beyond global probabilities alone, distinguishing CG-HAF from detector-only count-to-grade rules and hidden-feature fusion without implying that global--local acne modeling is itself unprecedented.

\subsection{Reliability, Evaluation, and Research Gap}
\label{sec:reliability-evaluation}

Reliable evaluation of an evidence-fusion system requires considering aggregate performance, class-specific behavior, ordinal error, calibration, and evidence boundaries together. The EADV Artificial Intelligence Task Force emphasizes transparency about data quality, intended use, and performance across diverse patient groups \cite{ref27}, and broader reviews identify acquisition sensitivity, limited generalizability, and insufficient real-world validation as recurring translation barriers \cite{ref28,ref29}. These concerns are acute for acne, where lesion visibility varies with skin tone: a dedicated review documents underrepresentation and image-standardization problems that can produce unequal performance across populations \cite{ref30}, and a systematic review of cosmetic-dermatology AI reports substantial variation in datasets and evaluation practice, so metrics must be read within their protocol rather than as protocol-independent rankings \cite{ref31}. Acne detector studies commonly report precision, recall, F1, and mAP \cite{ref19,ref20}, but a four-grade severity model additionally requires metrics sensitive to class imbalance and ordered error distance; class-specific sensitivity complements aggregate agreement because a candidate may improve an ordinal statistic while losing reliability for minority high-severity cases. Repeated stochastic training with probability aggregation can reduce dependence on a favorable initialization, and calibration analysis assesses whether confidence aligns with correctness, though these practices improve internal auditability without substituting for clinical validation. Related teledermatology work further argues that deployment-oriented systems should separate algorithmic filtering from downstream clinical decisions \cite{ref32}, and interpretability should rely on detector boxes and low-dimensional fusion coefficients rather than treating post hoc heatmaps as causal explanations.

Table~\ref{tab:related_work_comparison} positions representative studies against CG-HAF. Prior work has examined ordinal or burden-aware grading, lesion detection and counting, hybrid representation learning, and reliability-oriented evaluation, but generally in separate architectural configurations: detector-centric methods convert lesion evidence directly into a grade without a separate holistic representation, while ensemble and representation-fusion methods capture broader appearance but leave their lesion-burden contribution difficult to isolate at the decision layer. No prior framework combines global image-level severity probabilities with explicitly observable, detector-derived burden cues into a single, compact final-stage representation whose incremental value is isolated through progressive fusion-feature ablation, while also protecting sensitivity to Severe- and Very-Severe-class during model selection. CG-HAF addresses this specific gap without claiming precedence over prior global--local or lesion-aware acne methods.

\begin{center}
\centering
\captionof{table}{Representative acne-severity grading and lesion-aware approaches positioned relative to CG-HAF. Comparisons are methodological; differences in datasets, grading criteria, and evaluation protocols preclude direct ranking of performance.}
\label{tab:related_work_comparison}
\footnotesize
\setlength{\tabcolsep}{4pt}
\renewcommand{\arraystretch}{1.2}
\begin{tabularx}{\textwidth}{p{0.12\textwidth} p{0.15\textwidth} p{0.17\textwidth} X X}
\toprule
\textbf{Study} & \textbf{Dataset / task} & \textbf{Core approach} & \textbf{Evidence / fusion strategy} & \textbf{Distinction from CG-HAF} \\
\midrule
Zhang and Ma (2022)~\cite{ref15} & ACNE04; severity, count, localization & Joint classification--localization ensemble & Severity/count prediction with parallel localization outputs & Parallel outputs; detector burden not fused into final grading \\
Khairani and Kosala (2025)~\cite{ref16} & Acne detection; 4-class severity & YOLOv11 with count-to-grade mapping & Lesion counts converted to severity via count thresholds & Count-derived grading; no independent global probability stream \\
ClearFace (2025)~\cite{ref17} & Lesion detection/classification; IGA scoring & Detection, lesion-type classification, weighted scoring & Local detections/counts aggregated into an IGA score & Explicit scoring; no independent global probability stream \\
Zhang et al. (2024)~\cite{ref19} & ACNE04; lesion detection & Enhanced YOLOv7 detector & Multi-scale detector features for localization & Optimizes detection, not global--local grading fusion \\
Ilahi and Gunawan (2025)~\cite{ref20} & Multi-type lesion detection & YOLOv8 with optimizer-focused evaluation & Detector-internal boxes, classes, confidence & Detector optimization; no fusion with global severity \\
Nainggolan et al. (2025)~\cite{ref21} & Inflammatory lesion detection; skin-color analysis & YOLOv11 inflammatory detection & Local detection evaluated across skin-color scenarios & Detection fairness study, not final-stage grading fusion \\
\textbf{CG-HAF (present study)} & ACNE04; 4-class ordinal severity & ConvNeXtV2-Large + YOLOv12s structured fusion & Global severity probabilities fused with explicit detector-derived lesion-burden variables & Explicit final-stage fusion of global and structured local evidence \\
\bottomrule
\end{tabularx}
\end{center}
\FloatBarrier
\section{Materials}
\label{sec:materials}

\subsection{ACNE04 Dataset, Annotations, and Fixed Partition}
\label{subsec:dataset}

ACNE04 pairs facial images with global severity labels and lesion annotations for joint acne grading and counting \cite{ref01}: 1,457 images and 18,983 lesion boxes across four ordered grades (Mild, Moderate, Severe, Very Severe), split into fixed 1,049/116/292 train/validation/test partitions (Table~\ref{tab:acne04-composition}). The image was the unit of analysis; patient identifiers were unavailable, so subject-level independence could not be established. Validation supported checkpoint selection, detector-threshold selection, and meta-classifier fitting, while the test split provided internal confirmation -- though, having been inspected during prior development, it is not an untouched benchmark. Marked class imbalance at higher grades motivated class-balanced sampling and evaluation throughout.

\begin{center}
\centering
\captionof{table}{Fixed ACNE04 severity and lesion-annotation composition.}
\label{tab:acne04-composition}

\scriptsize
\setlength{\tabcolsep}{1.8pt}
\renewcommand{\arraystretch}{1.10}

\begin{tabular}{lrrrrrr}
\toprule
\textbf{Split} &
\textbf{Mild} &
\textbf{Mod.} &
\textbf{Sev.} &
\textbf{V. Sev.} &
\textbf{Images} &
\textbf{Boxes} \\
\midrule

Train
& 362 & 457 & 139 & 91
& 1049 & 13611 \\

Validation
& 32 & 59 & 13 & 12
& 116 & 1616 \\

Test
& 99 & 131 & 36 & 26
& 292 & 3756 \\

\midrule
\textbf{Total}
& \textbf{493}
& \textbf{647}
& \textbf{188}
& \textbf{129}
& \textbf{1457}
& \textbf{18983} \\

\bottomrule
\end{tabular}

\vspace{1pt}
\begin{minipage}{\columnwidth}
\scriptsize
\textit{Note:} Mod. = Moderate; Sev. = Severe;
V. Sev. = Very Severe; Boxes = lesion bounding-box annotations.
\end{minipage}

\end{center}

\subsection{PLSBRACNE01 External Evaluation Cohort}
\label{subsec:plsbracne01}

PLSBRACNE01 served as an independent cohort for evaluating cross-dataset transport under a different grading criterion \cite{ref05}: 200 subjects (three facial views each; 600 images), consensus-labeled across four grades (58/47/65/30 subjects, Grade I--IV). A SHA-256 audit confirmed no image overlap with ACNE04. The image package and inference manifest were fixed before label access, and dermatologist XML lesion annotations were reserved solely for the separate annotation-assisted diagnostic (Section~\ref{subsec:annotation-diagnostic}).

\section{Methods}
\label{sec:methods}

\subsection{Study Design and Evidence Scope}
\label{subsec:study_design}

This study retrospectively evaluates four-class ordinal acne severity grading, testing whether a compact, interpretable final representation can jointly preserve holistic image-level severity evidence and localized lesion-burden evidence. The pipeline comprises the fixed ACNE04 partition, a ConvNeXtV2-Large global branch, a YOLOv12s local lesion-evidence branch, probability aggregation across three independently trained global models, a 12-dimensional detector-to-grading representation, and a balanced logistic-regression meta-classifier fitted on validation data (Fig.~\ref{fig:cghaf-agentic-overview}).

\begin{center}
\centering
\IfFileExists{figures/Final_CG_HAF_Architecture.pdf}{%
  \includegraphics[width=\textwidth]{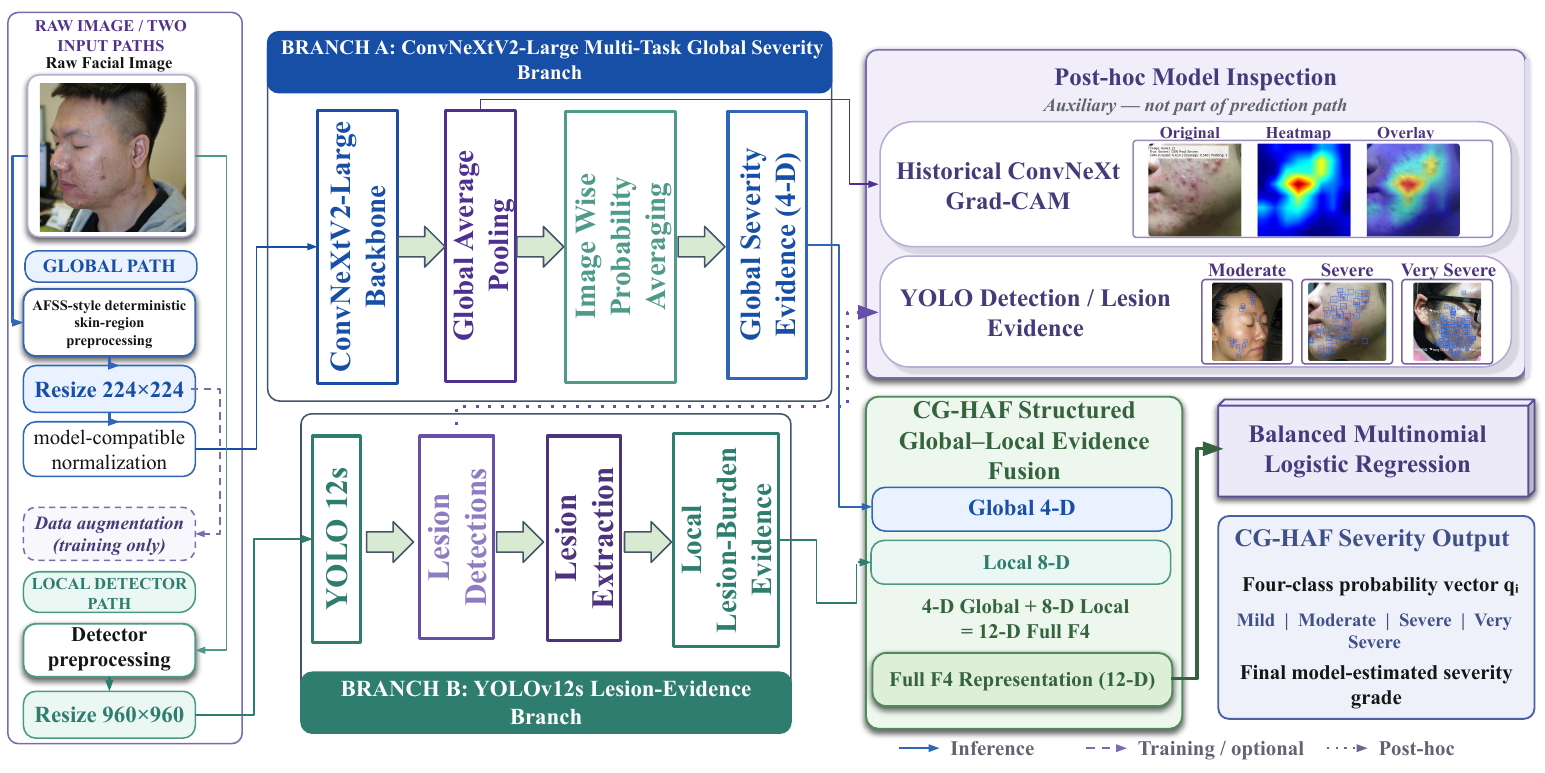}%
}{%
  \includegraphics[width=\textwidth]{figures/Figure_12.pdf}%
}
\captionof{figure}{Retained CG-HAF architecture. Three-seed ConvNeXtV2-Large probabilities provide 4-D global severity evidence, while YOLOv12s detections provide 8-D structured lesion-burden evidence. Their concatenation yields the 12-D F4 representation used by a validation-fitted balanced multinomial logistic regression classifier. The auxiliary count head is training-only, and Grad-CAM/YOLO visualizations are post hoc inspection components that do not affect predictions.}
\label{fig:cghaf-agentic-overview}
\end{center}

Four evidence tiers are analyzed: three-seed CG-HAF confirmation, controlled global-branch training ablations, progressive F0--F4 fusion-feature ablations, and a secondary count-conditioned candidate under a frozen high-severity non-degradation rule; these ACNE04 analyses are treated as internal confirmation given the historically inspected test partition. Two post-retention analyses use no ACNE04-style fitting: a frozen, label-blind cross-dataset/cross-criterion evaluation on PLSBRACNE01, and an annotation-assisted diagnostic using target labels and privileged lesion annotations (not interpreted as automatic external performance). No claim of prospective or clinical validation is made.

\subsection{Global ConvNeXtV2-Large Severity-Count Branch}
\label{subsec:global_branch}

\begin{center}
    \centering
    \includegraphics[width=0.9\columnwidth]{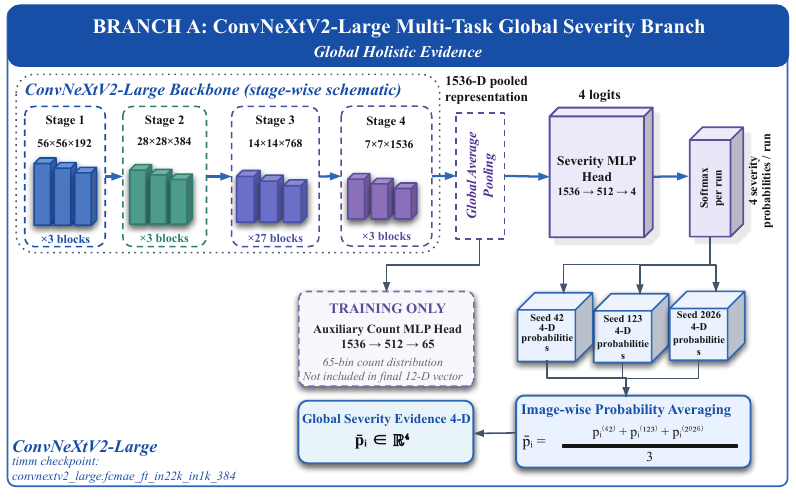}
    \captionof{figure}{Detailed architecture of the global ConvNeXtV2-Large severity-count branch, illustrating the backbone representation learning, four-class ordinal severity head, auxiliary 65-bin lesion-count distribution head, and Teacher-EMA consistency regularization pathway.}
    \label{fig:global-branch-detail}
\end{center}

The global branch uses ConvNeXtV2-Large \cite{ref33} (timm checkpoint \path{convnextv2_large.fcmae_ft_in22k_in1k_384}, fine-tuned at \(224\times224\)) with a four-class ordinal severity head and an auxiliary 65-bin lesion-count head (Fig.~\ref{fig:global-branch-detail}). Images are resized to \(224\times224\) after a deterministic skin-region masking step (YCrCb thresholding, morphological opening/dilation, boundary feathering, background suppression) -- a rule-based preprocessing step, not a learned segmentation network, evaluated as ablation A1 (Section~\ref{sec:controlled-ablations}). The auxiliary count head supplies a burden-learning signal and is aggregated into the four severity intervals for count-to-grade consistency but is not concatenated into the final fusion stack, keeping training-time count guidance separate from detector evidence used at fusion.

Training combines cross-entropy with Gaussian label distribution learning (\(\sigma=0.65\)) to model ambiguity between adjacent grades, auxiliary count-distribution and severity/count-agreement losses, and a Teacher-EMA consistency pathway between weak/strong augmented views \cite{ref35}:

\begin{equation}
\label{eq:global-objective}
\begin{aligned}
L ={}& 0.40\,L_{\mathrm{CE}} + 0.60\,L_{\mathrm{LDL}} + 0.30\,L_{\mathrm{count}}\\
&+ 0.40\,L_{\mathrm{aux}} + 0.20\,L_{\mathrm{KL\text{-}count}} + \lambda_{\mathrm{SSL}}L_{\mathrm{SSL}},
\end{aligned}
\end{equation}

with consistency weight capped at \(0.80\) and EMA decay ranging \(\sim\!0.995\)--\(0.9995\). Principal runs used seeds 42, 123, and 2026 (max 80 epochs, patience 18, batch size 4, gradient accumulation 8), AdamW with backbone/head learning rates \(6\times10^{-6}\)/\(6\times10^{-5}\), weight decay \(1.5\times10^{-2}\), five warm-up epochs, cosine decay, mixed precision, gradient clipping at norm 1.0, and class-balanced sampling.

\subsection{Local YOLOv12s Lesion-Evidence Branch}
\label{subsec:local_branch}

\begin{center}
    \centering
    \includegraphics[width=0.98\columnwidth]{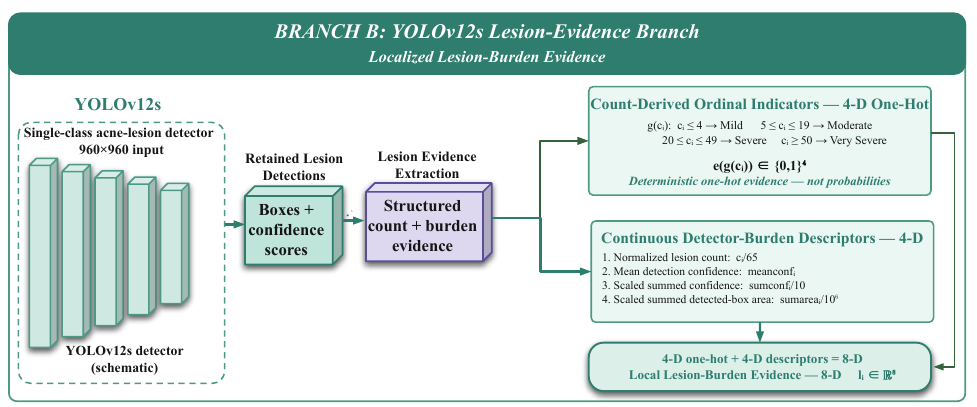}
    \captionof{figure}{Detailed architecture of the local YOLOv12s lesion-evidence branch, illustrating localized multi-scale acne lesion detection, bounding-box prediction, detection-confidence extraction, and count-to-grade severity mapping.}
    \label{fig:local-branch-detail}
\end{center}

The local branch uses YOLOv12s \cite{ref34}, a single-class acne-lesion detector (initialized from yolo12s.pt) trained at \(960\times960\) for 120 epochs (batch size 8, patience 30, SGD with learning rate 0.01, momentum 0.937, weight decay \(5\times10^{-4}\), 3-epoch warm-up, cosine schedule, mixed precision), preserving spatially localized lesion evidence independent of the global branch (Fig.~\ref{fig:local-branch-detail}). A single-class setting was used because fusion depends on total lesion burden, not lesion type. The final feature export used confidence threshold \(0.10\), IoU \(0.50\), and up to 300 detections per image (settings chosen via validation lesion-count error), retaining per-image predicted count \(c_i\), mean and summed detection confidence, and summed detected area (zero-valued if no detections). The predicted count maps to an ordered category via the Hayashi-aligned thresholds

\begin{equation}
\label{eq:count-to-grade}
g(c) =
\begin{cases}
\text{Mild}, & c \leq 4;\\
\text{Moderate}, & 5 \leq c \leq 19;\\
\text{Severe}, & 20 \leq c \leq 49;\\
\text{Very Severe}, & c \geq 50.
\end{cases}
\end{equation}

The detector supplies local lesion-burden evidence but does not itself decide the final grade; the CG-HAF prediction combines these variables with global severity probabilities.

\subsection{Three-Seed Aggregation and 12-Dimensional CG-HAF Fusion}
\label{subsec:seed_aggregation}
\label{subsec:fusion}

The global branch was trained independently with seeds 42, 123, and 2026 under the same partition and configuration; validation/test rows were aligned by image stem and averaged (no voting) to give the aggregated global probability

\begin{equation}
\label{eq:seed-probability-aggregation}
\bar{p}_{i,k} = \frac{1}{3}\sum_{s\in\{42,123,2026\}} p_{i,k}^{(s)}.
\end{equation}

This averaged vector \(\bar{p}_i\) forms the global evidence; local evidence adds a one-hot count-derived severity category \(e(g(c_i))\), normalized lesion count, mean/summed detection confidence, and summed lesion area -- 12 features total, fixed before the locked F0--F4 evaluation:

\begin{equation}
\label{eq:cghaf-fusion}
\begin{aligned}
z_i = [&\,\bar{p}_i,\ e(g(c_i)),\ c_i/65,\ \mathrm{meanconf}_i,\\
       &\,\mathrm{sumconf}_i/10,\ \mathrm{sumarea}_i/10^6] \in \mathbb{R}^{12},\\
\hat{y}_i &= \arg\max_k P(y=k\mid z_i),\\
k &\in \{\text{Mild},\text{Moderate},\text{Severe},\text{Very Severe}\}.
\end{aligned}
\end{equation}

A multinomial logistic-regression meta-classifier was fitted on the 116-image validation set with balanced class weights (\(L_2\), \(C=1.0\), lbfgs, max 3,000 iterations, tolerance \(10^{-4}\), random state 4042) and applied unchanged to test features, with normalization constants fixed (not learned from test data). A low-capacity final layer was chosen so the hybrid representation, not a second high-capacity network, remains the primary object of evaluation.

\subsection{Controlled Ablations and Safety-Constrained Candidate Evaluation}
\label{sec:controlled-ablations}

Training-component and fusion-feature ablations were analyzed as separate families to avoid conflating representation-learning effects with detector fusion (Table~\ref{tab:controlled-ablation-design}). The training family (evaluated at the global probability-mean endpoint) removes one component from the reference configuration at a time: A0 is a vanilla cross-entropy-only baseline; A1 removes skin-region masking; A2 removes label distribution learning; A3 removes the count head; A4 removes Teacher-EMA consistency. The fusion family progressively adds detector evidence to a newly refitted balanced logistic regression: F0 (global probabilities only) through F4 (complete 12-D representation), each refit rather than reusing prior coefficients. A secondary \(384\)-pixel count-conditioned ConvNeXtV2 candidate was also evaluated, combined with the principal branch as
\[
0.625\,p_{\mathrm{original}} + 0.375\,p_{\mathrm{count\text{-}conditioned}},
\]
but was not adopted, since candidate selection required Severe-/Very-Severe recall to not degrade even when aggregate or calibration metrics improved.

\begin{center}
\centering
\captionof{table}{Controlled training and fusion ablation design.}
\label{tab:controlled-ablation-design}
\small
\begin{tabularx}{\textwidth}{l l X X}
\hline
Family & ID & Controlled change & Purpose \\
\hline
Training\textsuperscript{a} & Full & Reference branch: masking + CE/LDL + count guidance + Teacher-EMA & Reference configuration \\
 & A0 & Vanilla CE-only classifier & Effect of full training design vs.\ minimal baseline \\
 & A1 & Remove skin-region masking & Contribution of masking \\
 & A2 & Remove LDL & Contribution of ordinal label-distribution supervision \\
 & A3 & Remove count head/losses & Contribution of count guidance \\
 & A4 & Remove Teacher-EMA consistency & Contribution of EMA consistency regularization \\
\hline
Fusion\textsuperscript{b} & F0 & Global probabilities only & Reference stack without detector evidence \\
 & F1 & F0 + count-derived grade indicators & Value of count-derived ordinal evidence \\
 & F2 & F1 + normalized lesion count & Incremental value of continuous burden \\
 & F3 & F2 + mean/summed confidence & Incremental value of confidence evidence \\
 & F4 & F3 + normalized lesion area (complete 12-D) & Value of complete representation \\
\hline
Candidate\textsuperscript{c} & CC & Count-conditioned ConvNeXtV2, frozen 0.625/0.375 branch fusion & Whether added complexity preserves high-severity sensitivity \\
\hline
\end{tabularx}
\vspace{1pt}
\begin{minipage}{\textwidth}
\scriptsize
\textsuperscript{a}Endpoint: three-seed global probability mean. \textsuperscript{b}Endpoint: newly validation-fitted balanced logistic regression at each step. \textsuperscript{c}Endpoint: frozen evaluation rule, no post-hoc test-weight tuning.
\end{minipage}
\end{center}

\begin{center}
    \centering
    \includegraphics[width=0.78\columnwidth]{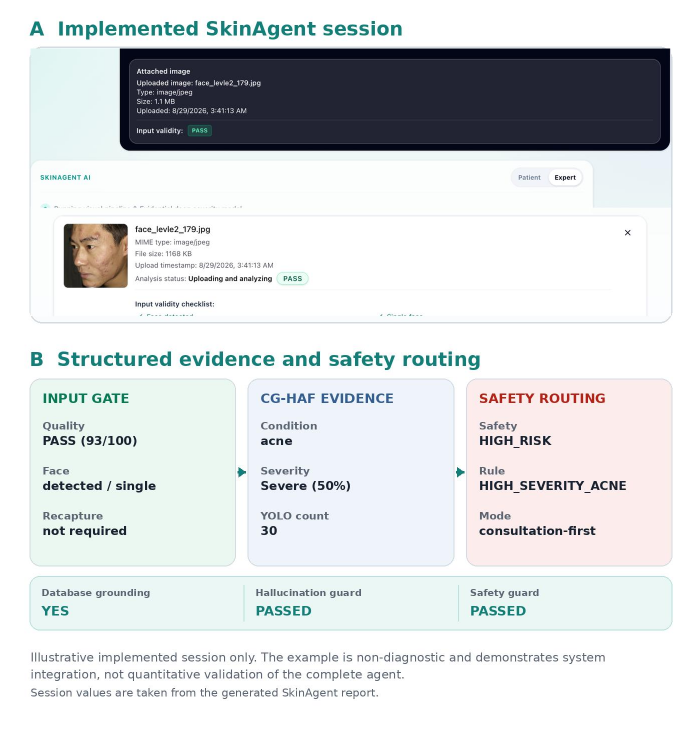}
    \captionof{figure}{Implemented SkinAgent application example showing input validation, CG-HAF-derived severity and lesion-burden evidence, and downstream safety-grounded routing. This example demonstrates system integration only and is not a quantitative or clinical validation of the complete agentic system.}
    \label{fig:skinagent-application}
\end{center}

\subsection{Integration within Agentic Skincare Support}
\label{sec:agentic-integration}

As a downstream illustration, CG-HAF also integrates as a structured, non-diagnostic visual-evidence module within a SkinAgent research prototype (Figure~\ref{fig:skinagent-application}): a validated input image yields a CG-HAF severity signal that a deterministic-guard, LLM-orchestrated pipeline consumes as fixed evidence (e.g., triggering consultation-first routing on high-severity predictions) without adapting CG-HAF's weights. This system context is secondary to, and outside the experimental scope of, the acne-grading methodology evaluated here.

\subsection{Evaluation, External Transport, and Criterion-Adaptation Diagnostic}
\label{subsec:evaluation}
\label{subsec:frozen-external}

Evaluation used accuracy, balanced accuracy, Macro-F1, class-wise precision/recall/F1, QWK, and grade MAE, with Severe/Very-Severe recall reported separately and calibration summarized via negative log-likelihood, Brier score, and 15-bin ECE \cite{ref36}. Paired exact McNemar tests, class-stratified bootstrap resampling (20,000 iterations, 95\% CIs), and Holm-adjusted family-wise error control quantified significance. Interpretability relied on YOLOv12s detection boxes, the semantically named 12-D fusion vector, and inspectable logistic-regression coefficients, treated as evidence rather than causal explanation.

The frozen ACNE04-trained system was then applied to PLSBRACNE01 without adaptation: all checkpoints, detector, and meta-classifiers were fixed before label access, with no retraining, refitting, or recalibration. All 600 views underwent label-blind inference, aggregated to one subject-level prediction each and evaluated (given PLSBRACNE01's different grading criterion) via accuracy, balanced accuracy, Macro-F1, QWK, grade MAE, and per-grade recall, with 20,000 bootstrap replicates for F4.

A separate privileged-information diagnostic then tested whether the transport gap reflected detection limitations or criterion misalignment: expert annotations were summarized into a ten-dimensional TYPE10 representation (mean/maximum counts of five lesion types) and a TOTAL2 burden-only comparator, compared against the locked ACNE04/Hayashi rule via nested five-fold subject-level cross-validation with bootstrap uncertainty and multi-seed stability checks. Using target-dataset labels, this diagnostic does not measure CG-HAF's automatic external performance.
\label{subsec:annotation-diagnostic}

\setcounter{equation}{4}
\FloatBarrier
\section{Experimental Results}
\label{sec:experimental-results}

\subsection{Evidence Integrity and Historical Reference}
\label{sec:evidence-integrity}

All internal ACNE04 comparisons used the same fixed 292-image test partition (Section~\ref{subsec:dataset}); because this partition had been inspected during CG-HAF's historical development, these results constitute internal confirmation rather than an untouched benchmark. A historical single-run CG-HAF, retained only as a development-trajectory reference, achieved 84.93\% accuracy (QWK 0.9042); the current three-seed F4 system scores numerically higher, but this specific comparison did not reach significance (McNemar \(p=0.210\)) and is not the paper's evidentiary basis. The controlled global--local branch and F0--F4 fusion ablations (Section~\ref{sec:ablation-studies}) provide the principal inferential evidence for the architecture. The complete historical/exploratory development ledger, including additional backbone and fusion variants, is reported in Supplementary Table~S1 for context rather than in the main text.

\subsection{Final Retained CG-HAF Performance}
\label{sec:final-retained-performance}

The retained three-seed, F4-fusion CG-HAF achieved 87.33\% accuracy (255/292 correct), balanced accuracy 0.8699, Macro-F1 0.8576, QWK 0.9173, and grade MAE 0.1301, with per-class recall ranging from 0.83 (Severe) to 0.90 (Mild). These values define the retained result used throughout the manuscript.

As shown in the confusion matrix (Figure~\ref{fig:retained-confusion}), the great majority of errors occurred between adjacent severity grades, consistent with the high QWK despite imperfect accuracy. Relative to the historical reference, gains were concentrated in the Moderate and Severe classes, while Very Severe recall declined slightly -- so the improvement is not uniform across all severity levels. Balanced accuracy and Macro-F1, rather than overall accuracy alone, better reflect performance given the small sizes of the high-severity classes.

\begin{center}
\centering
\includegraphics[width=0.5\columnwidth]{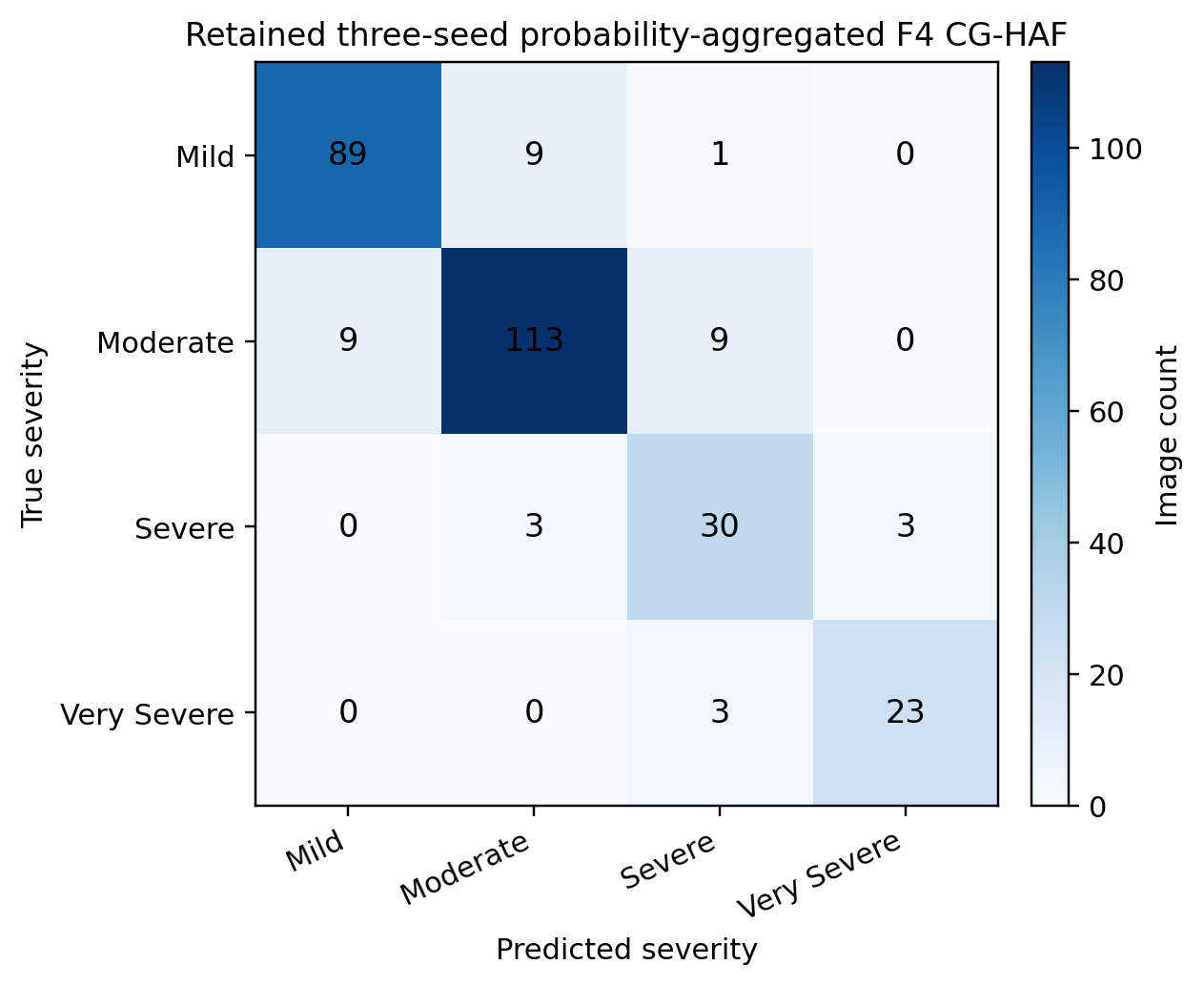}
\captionof{figure}{Confusion matrix of the retained three-seed probability-aggregated F4 CG-HAF on the fixed ACNE04 test split. Integer counts are shown; 36 of 37 errors occurred between adjacent severity grades.}
\label{fig:retained-confusion}
\end{center}


\subsection{Qualitative Comparison with Principal Baselines}
\label{subsec:qualitative-main-results}

Fig.~\ref{fig:qualitative-main-results} shows representative ACNE04 test images with predictions from the three principal aligned systems: the YOLOv12s count-only reference, the historical canonical CG-HAF, and the retained F4 CG-HAF.

\begin{center}
    \centering
    \includegraphics[width=1\textwidth]
    {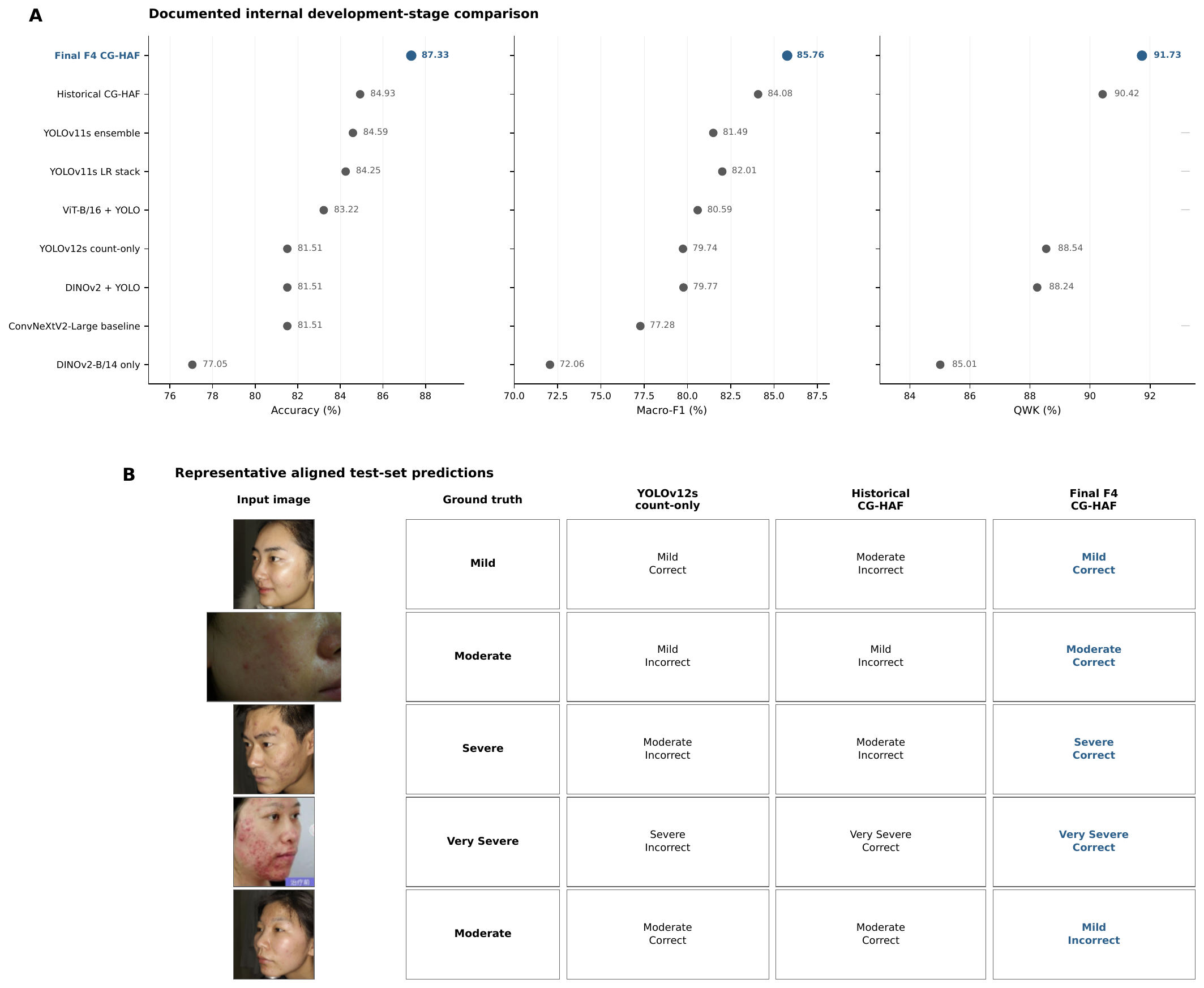}
    \captionof{figure}{Qualitative comparison of representative ACNE04 test
    cases using principal internal comparison systems with aligned
    per-image predictions. Each row shows the input image,
    ground-truth severity grade, and predictions from the YOLOv12s
    count-only reference, the historical canonical CG-HAF system,
    and the retained three-seed probability-aggregated F4 CG-HAF
    model. The examples illustrate image-level differences among
    the compared predictors and include a challenging case to avoid
    presenting only successful final-model predictions.}
    \label{fig:qualitative-main-results}
\end{center}

These cases illustrate that the aggregate metric differences reflect real image-level patterns: F4 often matches ground truth where comparison systems miss by an adjacent grade, though a retained challenging example shows the final model does not resolve every case. These observations are qualitative support, not independent evidence of statistical superiority.

\subsection{Multi-Seed Stability and Probability Aggregation}
\label{sec:multi-seed-results}

The three independently trained global-branch seeds gave somewhat different downstream F4 results (accuracy 84.9--86.3\%, Macro-F1 0.822--0.845, QWK 0.902--0.913), averaging 85.50\% \(\pm\) 0.71 pp accuracy across seeds -- this quantifies stochastic training variation but is not the final ensemble output.

Instead, the four class probabilities from the three runs were averaged image-wise before refitting the F4 meta-classifier. This probability-aggregated predictor reached 87.33\% accuracy, 0.8576 Macro-F1, and 0.9173 QWK -- exceeding both the seed-level mean and every individual seed, so it is reported as a distinct predictor rather than a ``mean accuracy.'' High-severity recall also varied by seed (Severe: 0.72--0.81; Very Severe: 0.85--0.92), and the aggregated predictor (0.8333, 0.8846) did not simply match the best individual seed for each class -- aggregation combines the underlying probability fields rather than selecting a winner. No formal variance-reduction analysis was conducted.

\begin{center}
\centering
\captionof{table}{Seed-level stability and probability-level aggregation.}
\label{tab:seed-stability}
\small
\resizebox{\textwidth}{!}{%
\begin{tabular}{lllll}
\hline
\textbf{Predictor} & \textbf{Accuracy} & \textbf{Macro-F1} & \textbf{QWK} & \textbf{Notes} \\
\hline
Seed 42 & 84.93\% & 0.8221 & 0.9018 & Independent F4 stack \\
Seed 123 & 86.30\% & 0.8441 & 0.9128 & Independent F4 stack \\
Seed 2026 & 85.27\% & 0.8445 & 0.9064 & Independent F4 stack \\
Seed mean \(\pm\) SD & 85.50\% \(\pm\) 0.71 pp & 0.8369 \(\pm\) 0.0128 & 0.9070 \(\pm\) 0.0055 & Arithmetic run-level summary \\
Probability-aggregated F4 & 87.33\% & 0.8576 & 0.9173 & Final image-wise probability aggregation \\
\hline
\end{tabular}
}
\vspace{0.35em}
\begin{minipage}{0.98\textwidth}
\footnotesize The probability-aggregated predictor is a separate inference rule and is not an error-barred fourth stochastic run.
\end{minipage}
\end{center}

\begin{center}
\centering
\includegraphics[width=0.5\columnwidth]{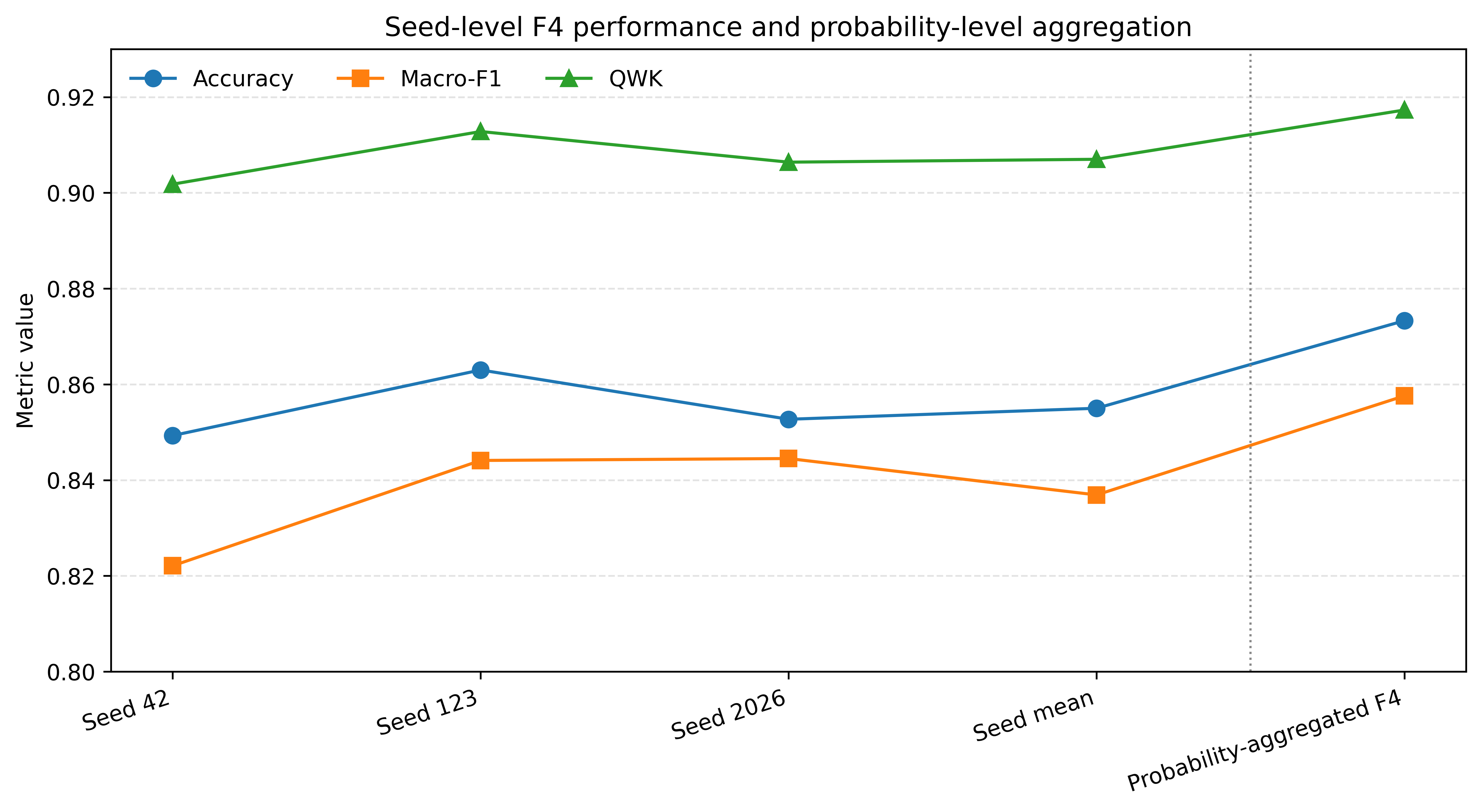}
\captionof{figure}{Seed-level F4 stacking performance and the distinct probability-aggregated F4 predictor. The seed mean is a descriptive run-level summary; the aggregated predictor has no run-level error bar.}
\label{fig:seed-aggregation}
\end{center}

\subsection{Safety-Constrained Candidate Evaluation}
\label{sec:safety-candidate-results}

The secondary count-conditioned branch and its frozen 0.625/0.375 combination with the retained branch were evaluated under the safety-constrained rule (Section~\ref{sec:controlled-ablations}): the count-conditioned candidate alone underperformed the retained system, and while the frozen dual-stack fusion matched its accuracy (87.33\%) and improved QWK, grade MAE, and calibration, it did so at the cost of a meaningful drop in Severe-class recall (0.8333 to 0.7500) with no net gain in correct predictions (McNemar \(p=1.000\)). Because it reduced Severe-class sensitivity without improving accuracy, the frozen dual stack failed the high-severity non-degradation gate and was not retained; the original-branch three-seed F4 remained the final CG-HAF predictor, reflecting model selection under the frozen internal protocol rather than clinical safety validation.

\begin{center}
\centering
\captionof{table}{Safety-constrained comparison of retained and secondary three-seed candidate systems.}
\label{tab:safety-candidate}
\small
\resizebox{\textwidth}{!}{%
\begin{tabular}{lllllllll}
\hline
\textbf{System} & \textbf{Accuracy} & \textbf{Macro-F1} & \textbf{QWK} & \textbf{Grade MAE} & \textbf{ECE} & \textbf{Severe recall} & \textbf{Very Severe recall} & \textbf{Decision} \\
\hline
Original-branch F4 & 87.33\% & 0.8576 & 0.9173 & 0.1301 & 0.1453 & 0.8333 & 0.8846 & Retained \\
Count-conditioned F4 & 86.99\% & 0.8557 & 0.9214 & 0.1301 & 0.1363 & 0.8056 & 0.8846 & Secondary \\
Frozen 0.625/0.375 dual stack & 87.33\% & 0.8486 & 0.9255 & 0.1267 & 0.1354 & 0.7500 & 0.9231 & Rejected \\
\hline
\end{tabular}
}
\vspace{0.35em}
\begin{minipage}{0.98\textwidth}
\footnotesize Retained-versus-dual paired correctness: 9 retained-only correct predictions, 9 dual-only correct predictions, and exact McNemar \(p = 1.000\). The dual stack was rejected because it provided no gain in accuracy and reduced Severe-class recall.
\end{minipage}
\end{center}

\subsection{Calibration, Error Patterns, and Interpretability}
\label{sec:calibration-interpretability}

The retained F4 predictor's calibration metrics (NLL 0.4750, Brier 0.2528, ECE 0.1453) are descriptive confidence-quality measures showing residual, nonzero calibration error rather than full calibration; as noted in Section~\ref{sec:safety-candidate-results}, better calibration alone did not justify selecting a candidate that lost Severe-class recall.

Consistent with the confusion matrix, errors were overwhelmingly between adjacent grades, with only one wider (Mild-to-Severe) error and no available subgroup analysis to attribute errors to morphology, skin tone, or other factors.

Interpretability was examined at three levels (Figure~\ref{fig:interpretability}): qualitative Grad-CAM and YOLOv12s visualizations (generated on the historical single-run branch, not the three-seed ensemble, and labeled accordingly) concentrate over acne-bearing regions for representative Severe and Very Severe cases; and the fitted F4 logistic-regression coefficients show that global-probability features carry the largest, consistently positive weight for their corresponding class, while count-derived and continuous detector features contribute class-dependent positive and negative weights. Because feature blocks are on different scales, these coefficients describe the fitted decision surface descriptively and are not a causal importance ranking.

\begin{center}
\centering
\includegraphics[width=0.8\textwidth]{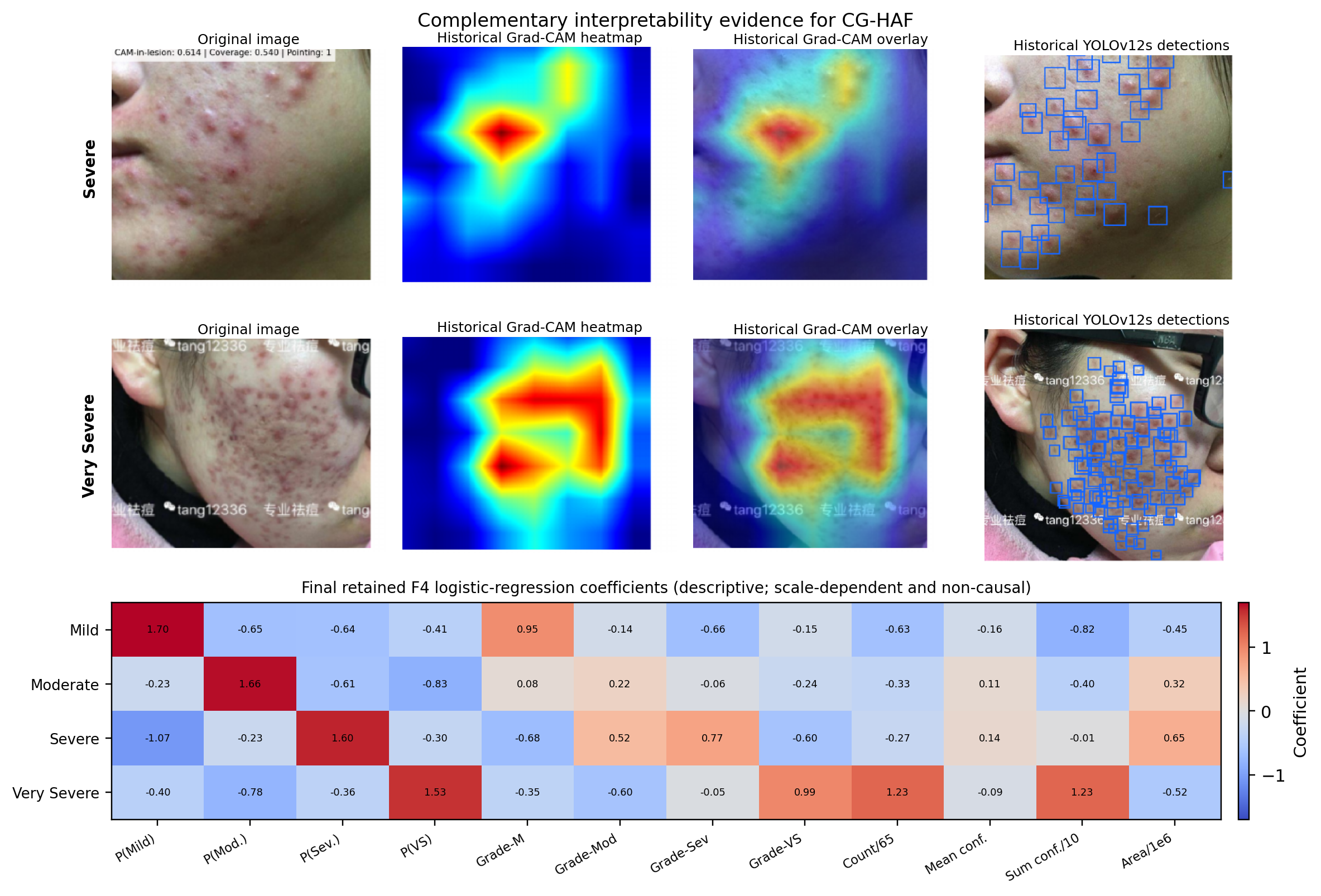}
\captionof{figure}{Complementary interpretability evidence for CG-HAF. The upper panels show historical qualitative ConvNeXtV2 Grad-CAM and YOLOv12s examples for aligned Severe (\texttt{levle3\_21}) and Very Severe (\texttt{levle3\_16}) images; they are not three-seed ensemble attributions. The lower panel shows coefficients from the final retained validation-fitted F4 logistic-regression stack. Coefficients are descriptive, scale-dependent, and non-causal.}
\label{fig:interpretability}
\end{center}

\subsection{Frozen Cross-Criterion Performance on PLSBRACNE01}
\label{subsec:external-results}

PLSBRACNE01 was a deliberately challenging transport setting, since both image distribution and grading criteria differed from those of ACNE04. Under this fully frozen setting, subject-level accuracy was modest for all three configurations (35--40\%), but F4 gave the strongest balanced accuracy, Macro-F1, QWK, and grade MAE among them (95\% bootstrap CI for F4 accuracy: 32.5--45.5\%). These results indicate limited zero-shot portability of the retained ACNE04 decision rule under simultaneous dataset and criterion shift, not a like-for-like replication.

\begin{center}
\centering
\captionof{table}{Frozen subject-level cross-criterion evaluation on
PLSBRACNE01.}
\label{tab:pls-frozen}
\small
\begin{tabular}{lccccc}
\hline
Model & Acc. & BAcc. & F1 & QWK & MAE \\
\hline
Global & 35.0 & 0.3346 & 0.3301 & 0.4390 & 0.770 \\
Local  & \textbf{40.0} & 0.3532 & 0.3362 & 0.4367 & 0.730 \\
F4     & 39.0 & \textbf{0.3554} & \textbf{0.3459} &
\textbf{0.4651} & \textbf{0.715} \\
\hline
\end{tabular}
\end{center}

\subsection{Annotation-Assisted Criterion-Adaptation Analysis}
\label{subsec:annotation-results}

The annotation-assisted analysis showed the transport gap could not be attributed solely to automatic detector error: accuracy on expert-derived lesion burden rose steadily from the locked Hayashi count rule (43.5\%) to a criterion-adapted total-burden model (58.5\%) to models using lesion-type composition (72--74\%), with the primary TYPE10 ordinal model performing best (74.0\% accuracy, Macro-F1 0.7364, QWK 0.8726; 95\% bootstrap CI 68.0--79.5\%; stable at \(73.1\%\pm1.64\) pp across five outer-partition seeds).

These results are an annotation-assisted diagnostic, kept separate from CG-HAF's automatic external performance; they indicate that criterion-specific mapping and lesion-type composition captured target-severity information that the fixed ACNE04 count-to-grade rule did not.

\begin{center}
\centering
\captionof{table}{Annotation-assisted cross-criterion diagnostic on
PLSBRACNE01. Expert lesion evidence is used in this analysis.}
\label{tab:pls-annotation}
\small
\begin{tabular}{lcccc}
\hline
Method & Acc. & Macro-F1 & QWK & MAE \\
\hline
Hayashi rule & 43.5 & 0.3629 & 0.5344 & 0.605 \\
TOTAL2 ordinal & 58.5 & 0.5874 & 0.7312 & 0.475 \\
TYPE10 multinomial & 72.0 & 0.7227 & 0.8614 & 0.290 \\
TYPE10 ordinal & \textbf{74.0} & \textbf{0.7364} &
\textbf{0.8726} & \textbf{0.270} \\
\hline
\end{tabular}
\end{center}


\section{Ablation Studies}
\label{sec:ablation-studies}

Three complementary ablation analyses were conducted to isolate
different sources of performance within CG-HAF. First, a controlled
global--local branch ablation examined whether the retained performance
could be reproduced using either evidence stream alone. Second, the
F0--F4 fusion-feature analysis progressively introduced structured
detector-derived evidence while holding the averaged global probabilities
and detector export fixed. Third, the A0--A4 training-component analysis
examined selected components of the global image branch at the
probability-mean endpoint. These analyses address distinct methodological
questions and are therefore reported separately rather than interpreted
as a single ablation hierarchy.


\subsection{Global--Local Branch Ablation}
\label{sec:branch-ablation-results}

To isolate the contribution of each evidence stream, three configurations were evaluated on the same fixed validation and test partitions: Global-only (the four three-seed-averaged severity probabilities), Local-only (the eight structured detector-derived variables), and the Full CG-HAF representation combining both. A separate validation-fitted balanced logistic-regression meta-classifier was fitted for each configuration under an identical protocol.

\begin{center}
\centering
\captionof{table}{Controlled global--local branch ablation on the fixed ACNE04
test partition.}
\label{tab:branch-ablation}
\footnotesize
\resizebox{\textwidth}{!}{%
\begin{tabular}{lllllllll}
\hline
\textbf{Configuration} &
\textbf{Dim.} &
\textbf{Accuracy} &
\textbf{Balanced Acc.} &
\textbf{Macro-F1} &
\textbf{QWK} &
\textbf{Grade MAE} &
\textbf{Severe recall} &
\textbf{Very Severe recall} \\
\hline
Global-only &
4 &
81.16\% &
0.7747 &
0.7658 &
0.8706 &
0.1952 &
0.6389 &
0.7692 \\

Local-only &
8 &
80.48\% &
0.7906 &
0.7820 &
0.8853 &
0.1952 &
0.6389 &
0.8462 \\

\textbf{Full CG-HAF} &
12 &
\textbf{87.33\%} &
\textbf{0.8699} &
\textbf{0.8576} &
\textbf{0.9173} &
\textbf{0.1301} &
\textbf{0.8333} &
\textbf{0.8846} \\
\hline
\end{tabular}
}
\vspace{0.35em}
\begin{minipage}{0.98\textwidth}
\footnotesize
Global-only retains the four averaged image-level severity
probabilities, Local-only retains the eight structured detector-derived
variables, and Full CG-HAF combines both streams in the complete
12-dimensional representation.
\end{minipage}
\end{center}

The complete representation outperformed both single-stream configurations across every reported metric, improving accuracy by 6.16 percentage points over Global-only and 6.85 points over Local-only, with corresponding gains in Macro-F1, QWK, and Severe-class recall. Neither single stream dominated the other: Global-only yielded marginally higher accuracy, whereas Local-only achieved higher balanced accuracy, Macro-F1, QWK, and Very Severe-class recall, indicating complementary rather than redundant evidence. Holm-adjusted exact McNemar tests confirmed that removing either stream produced a significant change in predictive correctness (Full vs.\ Global-only, \(p=0.000554\); Full vs.\ Local-only, \(p=0.005515\)), supporting the value of fusing holistic severity evidence with explicit lesion-burden evidence, though this does not establish the causal or clinical importance of any individual feature.


\subsection{Qualitative Global--Local Model Comparison}
\label{subsec:qualitative_comparison}

To complement the quantitative branch ablation (Section~\ref{sec:branch-ablation-results}), Fig.~\ref{fig:qualitative_comparison} presents five illustrative ACNE04 test cases spanning distinct prediction patterns across the Global-only, Local-only, and Full CG-HAF configurations, alongside the dataset-wide correctness-pattern frequencies in Panel~B.

\begin{center}
    \centering
    \includegraphics[width=1\textwidth]
    {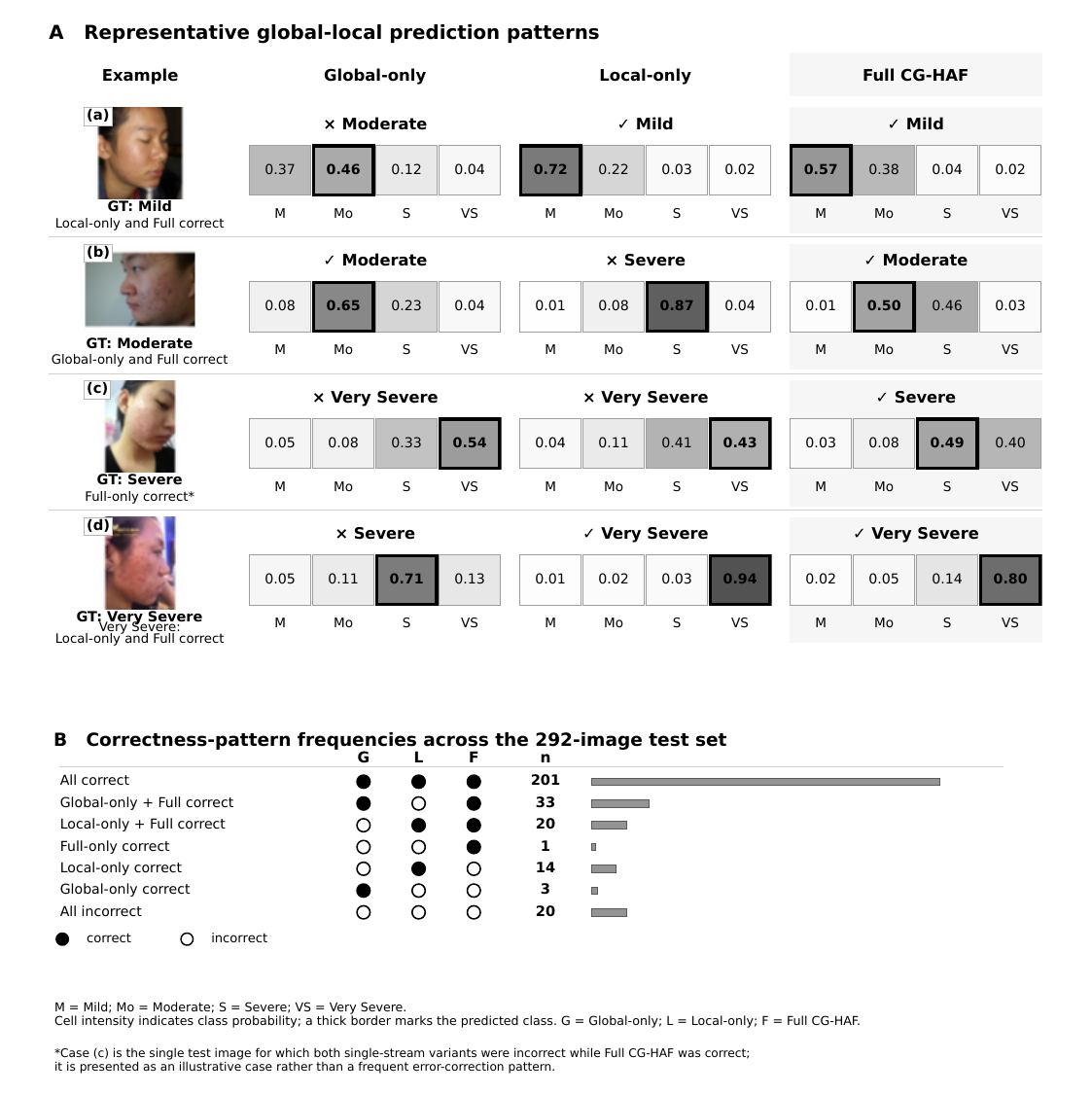}
    \captionof{figure}{Qualitative analysis of global--local prediction behavior on the ACNE04 test set. Panel~A shows five illustrative test images with ground-truth grades and the four-class predictive distributions from the Global-only, Local-only, and Full CG-HAF configurations (cell intensity = class probability; thick outline = predicted class), including complementary single-stream successes, one Full-only-correct case, and a fusion failure case. Panel~B reports correctness-pattern frequencies across the full 292-image test set. M/Mo/S/VS: Mild/Moderate/Severe/Very Severe; G/L/F: Global-only/Local-only/Full CG-HAF.}
    \label{fig:qualitative_comparison}
\end{center}

The selected cases illustrate several forms of complementarity: in some images only the local stream recovers the correct grade where the global stream errs, and vice versa, while one boundary case shows both single-stream variants converging on an incorrect Very Severe prediction that fusion corrects to Severe. A retained failure case shows that fusion does not always inherit a correct single-stream prediction. Panel~B situates these examples within the full test set: all three configurations agreed and were correct on 201 of 292 images, Full agreed with whichever single stream was correct in the remaining discordant cases far more often than it was uniquely correct, and the Full-only-correct pattern occurred just once. This supports complementary behavior between the two evidence streams while confirming that fusion does not eliminate all image-level errors.


\subsection{Progressive Fusion-Feature Ablation}
\label{sec:fusion-ablation-results}

The progressive F0--F4 analysis evaluated the detector-to-grading representation while holding the averaged global probabilities and detector export fixed. RAW\_CNN (direct argmax on the three-seed mean CNN probabilities) and F0 (a logistic-regression meta-classifier on the same probabilities) both achieved 81.16\% accuracy but are reported separately since they use different decision rules; F0 was modestly stronger on Macro-F1, QWK, and high-severity recall.

The largest single gain occurred from F0 to F1, when count-derived severity indicators were added (accuracy 81.16\%\(\to\)85.27\%, Macro-F1 0.7658\(\to\)0.8202, QWK 0.8706\(\to\)0.9053, Severe recall 0.6389\(\to\)0.7222); subsequent additions of continuous count (F2), detection confidence (F3), and lesion area (F4) yielded smaller, monotonic improvements, reaching 87.33\% accuracy and 0.8333/0.8846 Severe/Very-Severe recall at F4. The confirmatory F4-versus-F0 comparison showed a gain of 6.16 accuracy points, 0.0918 Macro-F1, 0.0467 QWK, and 0.1944 Severe recall, remaining significant after Holm correction (\(p=0.001109\)). This supports the complete structured representation over global-probability-only stacking, though the intermediate F1--F3 increments are not individually validated as significant after multiplicity correction and are presented as a decomposition rather than separate contributions.

\begin{center}
\centering
\includegraphics[width=0.7\columnwidth]{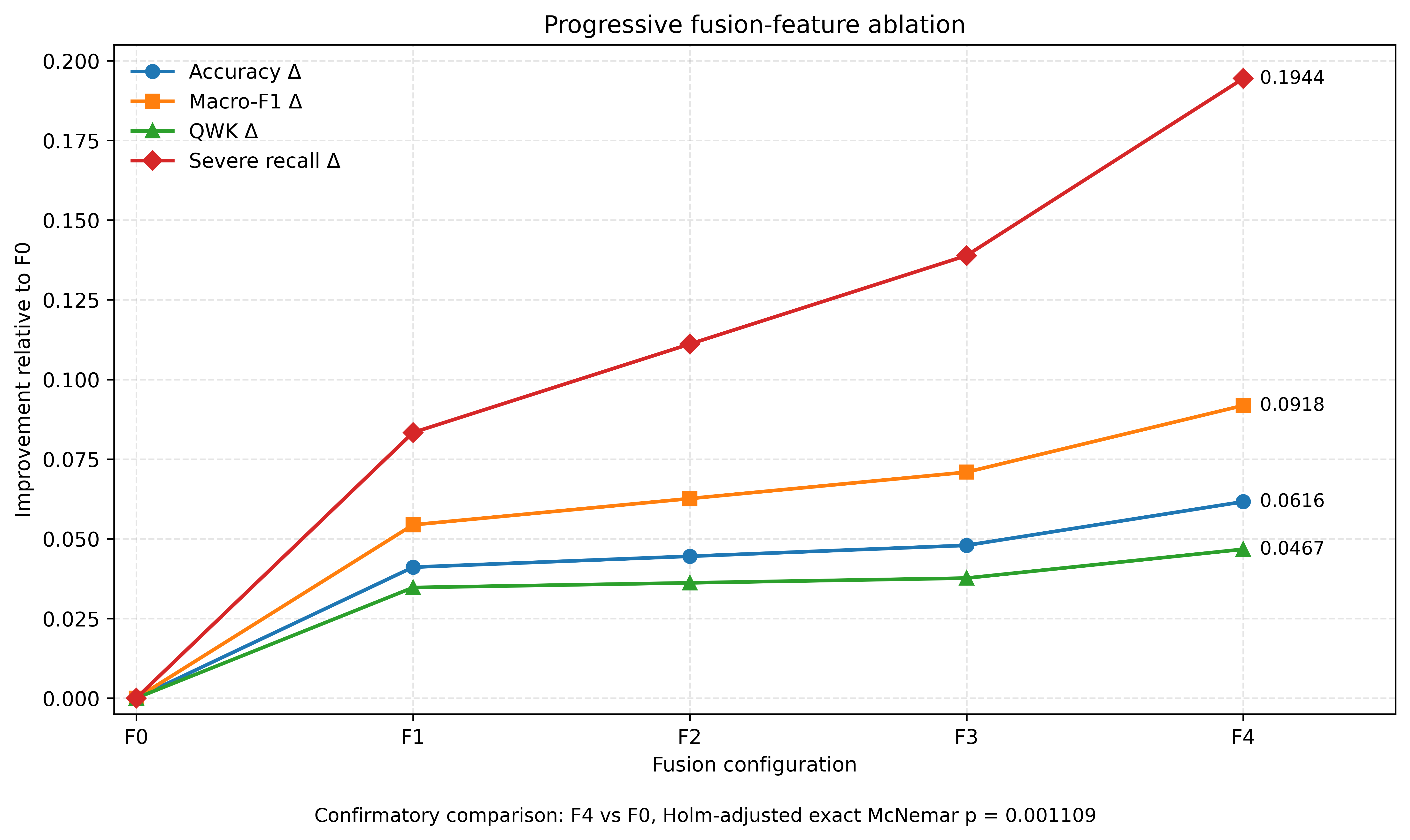}
\captionof{figure}{Progressive F0--F4 fusion-feature ablation shown as improvement
relative to F0. The confirmatory comparison concerns the complete F4
representation versus F0 (Holm-adjusted exact McNemar
\(p=0.001109\)); intermediate numerical increments are not interpreted
as independently significant.}
\label{fig:fusion-ablation}
\end{center}

\begin{center}
\centering
\captionof{table}{Progressive decomposition of the CG-HAF detector-to-grading
representation.}
\label{tab:fusion-decomposition}
\small
\resizebox{\textwidth}{!}{%
\begin{tabular}{lllllllll}
\hline
\textbf{Variant} &
\textbf{Feature composition} &
\textbf{Dim.} &
\textbf{Accuracy} &
\textbf{Macro-F1} &
\textbf{QWK} &
\textbf{Grade MAE} &
\textbf{Severe recall} &
\textbf{Very Severe recall} \\
\hline
RAW\_CNN & Direct three-seed mean CNN & 4 & 81.16\% & 0.7586 &
0.8689 & 0.1952 & 0.6111 & 0.7308 \\
F0 & Averaged global probabilities & 4 & 81.16\% & 0.7658 &
0.8706 & 0.1952 & 0.6389 & 0.7692 \\
F1 & F0 + count-derived grade & 8 & 85.27\% & 0.8202 &
0.9053 & 0.1507 & 0.7222 & 0.8462 \\
F2 & F1 + continuous count & 9 & 85.62\% & 0.8284 &
0.9068 & 0.1473 & 0.7500 & 0.8462 \\
F3 & F2 + confidence evidence & 11 & 85.96\% & 0.8367 &
0.9083 & 0.1438 & 0.7778 & 0.8462 \\
F4 & F3 + lesion area; complete 12-D & 12 & 87.33\% & 0.8576 &
0.9173 & 0.1301 & 0.8333 & 0.8846 \\
\hline
\end{tabular}
}
\vspace{0.35em}
\begin{minipage}{0.98\textwidth}
\footnotesize
Primary confirmatory comparison: F4 versus F0 yielded \(+6.16\)
percentage points in accuracy, \(+0.0918\) in Macro-F1, \(+0.0467\)
in QWK, and \(+0.1944\) in Severe-class recall; Holm-adjusted exact
McNemar \(p=0.001109\).
\end{minipage}
\end{center}


\subsection{Training-Component Ablation}
\label{sec:training-ablation-results}

Training-component ablation, evaluated at the global CNN probability-mean endpoint independent of the F0--F4 fusion analysis, showed that the complete training recipe did not uniformly dominate every reduced configuration: removing skin-region masking or using vanilla cross-entropy alone left several metrics essentially unchanged or slightly improved. Removing label distribution learning (A2), however, produced the clearest and most consistent deterioration (accuracy 81.16\%\(\to\)78.42\%; Severe recall 0.6111\(\to\)0.5000), while removing count guidance or Teacher-EMA consistency had smaller, less consistent effects.

No Full-versus-ablation comparison reached significance on the Holm-adjusted McNemar test, but a paired bootstrap gave A2 clear component-level support, with 95\% confidence intervals for balanced accuracy, Macro-F1, QWK, and Severe recall all excluding zero. Label distribution learning is therefore the one training component with independently supported evidence of benefit; the others are best regarded as inconclusive.

\begin{center}
\centering
\captionof{table}{Controlled training-component ablation using three-seed
probability-mean global predictions.}
\label{tab:training-ablation}
\small
\resizebox{\textwidth}{!}{%
\begin{tabular}{llllllll}
\hline
\textbf{Configuration} &
\textbf{Accuracy} &
\textbf{Balanced Acc.} &
\textbf{Macro-F1} &
\textbf{QWK} &
\textbf{Grade MAE} &
\textbf{Severe recall} &
\textbf{Very Severe recall} \\
\hline
Full image branch & 81.16\% & 0.7619 & 0.7586 & 0.8689 &
0.1952 & 0.6111 & 0.7308 \\
A0: vanilla CE-only & 80.14\% & 0.7774 & 0.7792 & 0.8756 &
0.1986 & 0.6667 & 0.8077 \\
A1: without skin-region masking & 81.51\% & 0.7922 & 0.7870 &
0.8884 & 0.1849 & 0.6111 & 0.8846 \\
A2: without LDL & 78.42\% & 0.7241 & 0.7247 & 0.8498 &
0.2226 & 0.5000 & 0.7308 \\
A3: without count guidance & 80.48\% & 0.7610 & 0.7593 &
0.8643 & 0.2021 & 0.5556 & 0.8077 \\
A4: without Teacher-EMA consistency & 79.79\% & 0.7518 &
0.7522 & 0.8583 & 0.2089 & 0.6111 & 0.7308 \\
\hline
\end{tabular}
}
\vspace{0.35em}
\begin{minipage}{0.98\textwidth}
\footnotesize
No Full-versus-ablation accuracy comparison remained significant after
Holm correction. For A2 minus Full, the bootstrap 95\% confidence
intervals excluded zero for balanced accuracy, Macro-F1, QWK, and
Severe-class recall.
\end{minipage}
\end{center}

\begin{center}
\centering
\includegraphics[width=0.8\columnwidth]{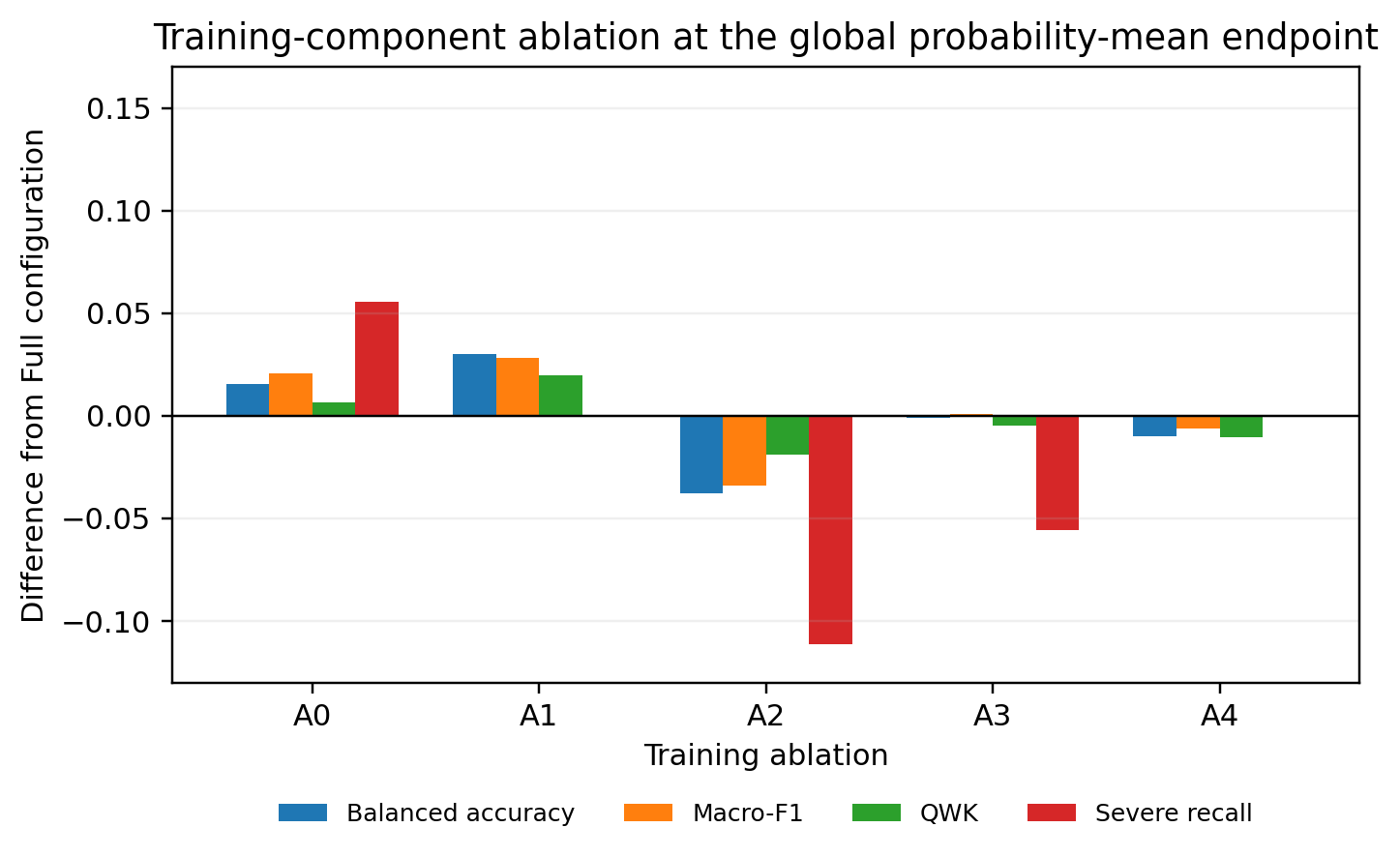}
\captionof{figure}{Differences between A0--A4 and the Full global image branch at
the probability-mean endpoint. The plot is descriptive; paired bootstrap
confidence intervals supporting the A2 finding are reported in the text
and are not inferred for all variants.}
\label{fig:training-ablation}
\end{center}
\FloatBarrier
\section{Discussion}
\label{sec:discussion}

\subsection{Principal Findings}
\label{sec:discussion-principal-findings}

The retained CG-HAF configuration was the three-seed probability-aggregated original-branch F4 predictor, which combined global severity probabilities with the complete detector-derived lesion-burden representation. On the fixed ACNE04 test partition, this internally confirmed system achieved 87.33\% accuracy, a Macro-F1 of 0.8576, a quadratic weighted kappa (QWK) of 0.9173, and a grade MAE of 0.1301. These values were numerically higher than the historical canonical single-run accuracy of 84.93\%, but the paired comparison between the historical and final predictions did not establish statistical superiority (McNemar \(p = 0.210\)). The historical difference should therefore be interpreted as an updated internal result, not as confirmatory evidence that the final predictor improved upon the earlier predictor.

The controlled fusion experiment provided more direct evidence for the architectural design. Relative to global-probability-only F0 stacking, the complete F4 representation increased accuracy by 6.16 percentage points, Macro-F1 by 0.0918, QWK by 0.0467, and Severe-class recall by 0.1944. The Holm-adjusted exact McNemar test yielded \(p = 0.001109\), supporting the complete structured lesion-evidence representation as the principal methodological contribution. This result does not establish that each successive feature addition from F1 through F4 was independently significant.

\subsection{Interpreting Global--Local Lesion-Burden Fusion}
\label{sec:discussion-global-local}

The gains from fusion are best explained by complementarity rather than redundancy: the global branch captures holistic appearance cues (erythema, texture, overall lesion distribution) without exposing how much of that signal comes from localized lesion burden, while the detector branch supplies exactly that missing burden signal as explicit count, confidence, and area variables. The largest single gain in the progressive fusion analysis (F0\(\to\)F1, Section~\ref{sec:fusion-ablation-results}) came from adding the coarse count-derived grade, indicating that this ordinal burden information was not already recoverable from the global probabilities; the remaining continuous descriptors then refine that coarse signal. Because the meta-classifier is a low-capacity logistic regression, this result speaks to the information content of the structured representation itself rather than to the added capacity of a larger fusion network, and the resulting coefficients, while inspectable, do not constitute proof of causal or clinical importance.

\subsection{Model Selection Under Stochasticity}
\label{sec:discussion-multiseed-selection}

Probability-level aggregation across three seeds was adopted because the training procedure remained sensitive to initialization; averaging class probabilities before fusion, rather than any single run, reduces this dependence without constituting a formal robustness guarantee. Among the training components examined in Section~\ref{sec:training-ablation-results}, only label distribution learning showed independently supported benefit, consistent with the ordered, boundary-ambiguous nature of acne severity labels -- the other components should be treated as unresolved rather than validated or refuted. The rejection of the count-conditioned dual-stack candidate (Section~\ref{sec:safety-candidate-results}) illustrates the value of a high-severity non-degradation criterion: selecting on QWK or calibration alone would have accepted a model with materially reduced Severe-class recall despite no net gain in correct predictions. This reflects safety-constrained selection within the internal protocol, not clinical validation.

\subsection{Calibration, Error Behavior, and Interpretability}
\label{sec:discussion-calibration-interpretability}

The retained predictor's residual calibration error (Section~\ref{sec:calibration-interpretability}) is a reminder that strong classification and ordinal agreement do not by themselves imply reliable confidence estimates, consistent with the rejected candidate's calibration/sensitivity trade-off noted above. Errors were overwhelmingly adjacent-grade rather than distributed across the full label space, consistent with an inherently ordinal task, though no subgroup analysis was available to attribute individual errors to skin tone, imaging conditions, or demographic factors. Interpretability was likewise kept at three inspectable levels -- attention visualizations, detector boxes, and fusion-layer coefficients -- none of which should be read as causal or clinically validating evidence, but together they make the pipeline considerably more auditable than a prediction derived from a single opaque embedding.

\subsection{Positioning Relative to Prior Acne-Grading Research}
\label{sec:discussion-prior-work}

CG-HAF builds on established ordinal-learning and lesion-aware grading work -- label distribution and grade-count supervision \cite{ref01}, global--local fusion \cite{ref04}, prior-knowledge and distillation-guided evidence \cite{ref10,ref05}, detector-centered counting \cite{ref06}, feature-feedback learning under heterogeneous grading standards \cite{ref11}, and lesion-level segmentation-based scoring \cite{ref12} -- but keeps the complementary evidence visible rather than latent: global probabilities and detector-derived burden variables remain explicit within a compact 12-D representation, evaluated progressively (F0--F4) and selected under a high-severity non-degradation rule rather than a single endpoint score. This supports a specific evidence-fusion contribution but not a state-of-the-art claim, since compared studies differ in data splits, domains, and evaluation protocols.

Table~\ref{tab:published_performance} summarizes reported performance from representative studies in citation order as context, not a controlled ranking, given these protocol differences.

\begin{center}
\centering
\captionof{table}{Reported performance of representative acne-severity grading
and lesion-aware studies, arranged in citation order.}
\label{tab:published_performance}

\footnotesize
\setlength{\tabcolsep}{2.5pt}
\renewcommand{\arraystretch}{1.14}

\begin{tabularx}{\columnwidth}{
@{}p{0.34\columnwidth}
p{0.31\columnwidth}
X@{}
}
\hline
\textbf{Study} &
\textbf{Dataset / task} &
\textbf{Reported result} \\
\hline

Wu et al. \cite{ref01} &
ACNE04; 4-grade severity &
Acc.: 84.11\% \\
\hline

KIEGLFN \cite{ref04} &
ACNE04 / PLSBRACNE01; grading &
Acc.: 84.52\% / 59.35\% \\
\hline

DED \cite{ref05} &
ACNE04 / PLSBRACNE01; grading &
Acc.: 86.06\% / 67.56\% \\
\hline

AcneGrader \cite{ref09} &
ACNE04; severity grading &
Acc.: 85.82\% \\
\hline

Prior-Knowledge Guided \cite{ref10} &
ACNE04 / PLSBRACNE01; grading &
Acc.: 85.27\% / 65.85\% \\
\hline

FF-PLL \cite{ref11} &
ACNE04 / ACNE-ECKH; grading &
Acc.: 87.33\% / 67.50\% \\
\hline

Zhang and Ma \cite{ref15} &
ACNE04; 3-grade severity$^{*}$ &
Acc.: 99.31\% \\
\hline

Khairani and Kosala \cite{ref16} &
ACNE04; severity + detection &
Acc.: 77.53\%;
mAP@50: 40.5\% \\
\hline

ClearFace \cite{ref17} &
Acne21; lesion + IGA grading &
Severity Acc.: NR \\
\hline

Zhang et al. \cite{ref19} &
ACNE04; lesion detection &
mAP: 83.7\% \\
\hline

Ilahi and Gunawan \cite{ref20} &
Skin90 + DermNet; detection &
mAP@0.5: 73.5\%;
Prec.: 82.3\% \\
\hline

Nainggolan et al. \cite{ref21} &
3,118 facial images; detection &
mAP: 88.2\%;
F1: 86.6\% \\
\hline

\textbf{CG-HAF (ours)} &
\textbf{ACNE04; 4-grade severity} &
\textbf{Acc.: 87.33\%;
F1: 0.8576;
QWK: 0.9173} \\
\hline

\textbf{TYPE10 ordinal (ours)} &
\textbf{PLSBRACNE01; grading} &
\textbf{Acc.: 74.0\%;
F1: 0.7364;
QWK: 0.8726} \\
\hline

\end{tabularx}

\vspace{2pt}
\begin{minipage}{\columnwidth}
\scriptsize
\textit{Note:} Results are reproduced according to the respective
studies and provide performance context rather than a controlled
ranking because grading criteria, datasets, class definitions, and
evaluation protocols differ. Detection mAP and lesion-level F1 are
not directly comparable with whole-image severity accuracy.
NR denotes that a directly comparable whole-image severity accuracy
was not verified.
$^{*}$Zhang and Ma reformulated ACNE04 into a three-class severity
setting (mild, moderate, and severe), so the reported accuracy is not
directly comparable with four-class CG-HAF.
$^{\dagger}$The 74.0\% PLSBRACNE01 result is an annotation-assisted
nested-cross-validation diagnostic using expert lesion annotations and
target-criterion labels; it is not automatic external CG-HAF performance.
\end{minipage}
\end{center}

\subsection{External Transport and Applications}
\label{subsec:external-discussion}
\label{subsec:applications}

The PLSBRACNE01 experiments address portability rather than the within-dataset fusion question: the frozen evaluation showed limited zero-shot transport when both the image distribution and grading criterion changed, while the annotation-assisted diagnostic indicates this gap owes to criterion misalignment as much as detection error, since performance improved once the criterion was adapted and lesion type made explicit -- though privileged expert annotations mean this does not reflect deployable performance, and the confounded domain/criterion shift precludes separating their individual effects. In application, CG-HAF integrates as a structured, non-diagnostic visual-evidence component within the SkinAgent prototype (Section~\ref{sec:agentic-integration}), where a model-estimated severity grade and lesion-count evidence can trigger safety routing (e.g., consultation-first flows for high-severity signals) independent of free-form language generation; this demonstrates implementation-level integration only, not end-to-end validation of recommendation quality, agent performance, or clinical safety, so broader deployment claims require separate system-level evaluation.
\FloatBarrier
\FloatBarrier

\section{Conclusion}
\label{sec:conclusion}

This study presented CG-HAF, an interpretable global--local framework for four-class ordinal acne severity grading that fuses holistic image-level severity probabilities from a ConvNeXtV2-Large branch with explicit lesion-burden evidence from a YOLOv12s detector in a compact, semantically defined 12-dimensional representation. On the fixed ACNE04 test partition, the retained three-seed F4 predictor achieved 87.33\% accuracy, a Macro-F1 of 0.8576, a QWK of 0.9173, and a grade MAE of 0.1301, improving on global-probability-only stacking by 6.16 accuracy points and 0.1944 Severe-class recall (Holm-adjusted McNemar \(p=0.001109\)). These results indicate that structured detector-derived lesion evidence adds decision-relevant information beyond global severity probabilities, positioning explicit detector-to-grading fusion as the framework's principal contribution.

At the same time, the frozen cross-dataset evaluation on PLSBRACNE01 showed limited zero-shot transfer under simultaneous dataset and criterion shift, and the annotation-assisted diagnostic suggests this gap reflects criterion misalignment as much as detection error -- a reminder that strong internal performance does not, by itself, establish external validity. Given the single inspected benchmark, unverified subject independence, and the absence of demographic and multicenter validation, CG-HAF remains a non-diagnostic research framework rather than a clinically validated tool. Future work should prioritize criterion-aware lesion-type evidence, external calibration, uncertainty-aware deferral, and prospective, subject-identified, multicenter evaluation before broader clinical or deployment claims can be made.
\FloatBarrier
\section*{CRediT authorship contribution statement}

Muhammad Muhtasim Shahriar: Conceptualization, Methodology, Software, Validation, Formal analysis, Investigation, Data curation, Visualization, Project administration, Writing -- original draft, Writing -- review \& editing.

Md. Naimur Asif Borno: Supervision, Validation, Visualization, Writing -- review \& editing.

Saad Aloteibi: Supervision, Project administration, Funding acquisition, Writing -- review \& editing.

Mohammad Ali Moni: Supervision, Project administration, Methodology, Resources, Writing -- review \& editing.

\section*{Declaration of competing interest}

The authors declare that they have no known competing financial interests or personal relationships that could have appeared to influence the work reported in this paper.

\section*{Data availability}

The ACNE04 dataset \cite{ref01} and the PLSBRACNE01 dataset \cite{ref04,ref05} used in this study are available from the sources cited in the manuscript, subject to the respective access conditions of their providers. No new dataset was created as part of this study.


\bibliographystyle{unsrtnat}
\bibliography{references}

\end{document}


\title[mode=title]{Supplementary Material: CG-HAF}
\maketitle
\section*{Supplementary Results}

\begin{table*}[t]
\centering
\caption{Supplementary Table S1. Complete documented internal development ledger for acne-severity experiments. Rows span historical thesis-stage model search, later DINOv2 exploration, secondary candidate evaluation, and the final retained three-seed CG-HAF. They are not a single controlled ablation family and should not be interpreted as a protocol-independent leaderboard.}
\label{tab:supp-internal-ledger}
\scriptsize
\begin{tabularx}{\textwidth}{p{3.2cm} X r r r p{2.4cm}}
\toprule
\textbf{Configuration} & \textbf{Short description} & \textbf{Accuracy} & \textbf{Macro-F1} & \textbf{QWK} & \textbf{Provenance / status} \\
\midrule
Final three-seed F4 CG-HAF & Probability-aggregated ConvNeXtV2 + YOLOv12s structured 12-D fusion & 87.33\% & 0.8576 & 0.9173 & Retained final \\
Frozen 0.625/0.375 dual stack & Original + count-conditioned branch probability fusion & 87.33\% & 0.8486 & 0.9255 & Rejected secondary candidate \\
Three-seed count-conditioned F4 & Count-conditioned ConvNeXtV2 branch + F4 stack & 86.99\% & 0.8557 & 0.9214 & Secondary candidate \\
Historical canonical CG-HAF & Historical ConvNeXtV2 + YOLOv12s hybrid reference & 84.93\% & 0.8408 & 0.9042 & Historical reference \\
Historical YOLOv11s ensemble variant & CNN severity probabilities + YOLOv11s detector-derived lesion evidence & 84.59\% & 0.8149 & -- & Historical exploratory \\
Historical YOLOv11s logistic stack & CNN probabilities + YOLOv11s count features + logistic regression & 84.25\% & 0.8201 & -- & Historical baseline \\
ViT-B/16 + YOLO stacking & ViT probabilities fused with detector evidence & 83.22\% & 0.8059 & -- & Historical exploratory \\
ConvNeXtV2 + ViT-B/16 + YOLO stacking & Dual-backbone hybrid + detector evidence & 82.88\% & 0.8058 & -- & Historical exploratory \\
ConvNeXtV2 + Swin-Tiny + YOLO stacking & Dual-backbone hybrid + detector evidence & 82.88\% & 0.8052 & -- & Historical exploratory \\
XGBoost hybrid & CNN probabilities + count features + XGBoost fusion & 82.88\% & 0.7756 & -- & Historical exploratory \\
Swin-Tiny + YOLO stacking & Swin probabilities + detector evidence & 82.53\% & 0.7975 & -- & Historical exploratory \\
DINOv2 + YOLOv12s & DINOv2-B/14 probabilities + YOLOv12s evidence & 81.51\% & 0.7977 & 0.8824 & Later exploratory run; not retained \\
YOLOv12s count-only & YOLOv12s count-derived severity only & 81.51\% & 0.7974 & 0.8854 & Local-only reference \\
ConvNeXtV2-Large baseline & Pure image-level ConvNeXtV2 classifier & 81.51\% & 0.7728 & -- & Historical global baseline \\
Historical CNN-only branch & Thesis-stage image-only branch; not current F0 & 81.16\% & 0.7931 & -- & Historical thesis-stage result \\
YOLOv11s count-only branch & YOLOv11s count evidence only & 80.48\% & 0.7221 & -- & Historical ablation \\
Swin-Tiny backbone-only & Pure transformer image-level baseline & 79.79\% & 0.7639 & -- & Historical baseline \\
ConvNeXtV2 + Swin-Tiny hybrid & Dual-backbone hybrid without detector stacking & 78.42\% & 0.7643 & -- & Historical exploratory \\
ConvNeXtV2 + ViT-B/16 hybrid & Dual-backbone hybrid without detector stacking & 77.74\% & 0.7462 & -- & Historical exploratory \\
5-fold CV ensemble + EMA + TTA & Training-intensive ConvNeXt ensemble baseline & 77.40\% & 0.7107 & -- & Historical exploratory \\
DINOv2-B/14 only & Standalone DINOv2 global severity branch & 77.05\% & 0.7206 & 0.8501 & Later exploratory run; not retained \\
ViT-B/16 backbone-only & Pure transformer image-level baseline & 75.34\% & 0.7355 & -- & Historical baseline \\
ResNet101d baseline & Conventional CNN image-level baseline & 71.58\% & 0.6693 & -- & Historical baseline \\
EfficientNet-B7 baseline & Conventional CNN image-level baseline & 63.70\% & 0.6234 & -- & Historical baseline \\
\bottomrule
\end{tabularx}
\vspace{0.25em}
\begin{minipage}{0.98\textwidth}
\footnotesize Notes: ``--'' denotes a metric not available in the documented source row and is not reconstructed. The historical thesis-stage CNN-only result is deliberately distinguished from the current RAW\_CNN and F0 definitions. The final 87.33\% result is produced by the three-seed probability-aggregated original-branch F4 predictor, not by the count-conditioned or frozen dual-stack candidates.
\end{minipage}
\end{table*}

\bibliographystyle{unsrtnat}
\bibliography{references}